\documentclass[12pt]{report}
\usepackage{pdfpages} 
\usepackage{anysize,fancyhdr,graphics}
\usepackage{csthesis}
\usepackage{longtable}
\usepackage[titletoc]{appendix}

\usepackage[T1]{fontenc}
\usepackage[utf8]{inputenc}
\usepackage{CJKutf8}
\usepackage{wrapfig}
\usepackage[font={scriptsize}]{caption}
\usepackage{tikz}
\usepackage{amsmath}

\usetikzlibrary{positioning,shapes,arrows,matrix,calc}

\newcommand{\brand}{}
\newcommand{\degrees}{\,^{\circ}\mathrm{C}}

\makeindex

\begin{document}
\setlength{\pdfpagewidth}{8.5in}
\setlength{\pdfpageheight}{11in}
\figurespagefalse
\tablespagefalse
\dedicationpagefalse
\quotationpagefalse
\otherlistpagefalse


\title{\large DORI:\\
       Distributed Outdoor Robotic Instruments \\[0.4cm]
       Hardware Design}
       
\author{Andrew Fuller}
\qualification{}
\submitdate{Spring 2013}
\copyrightyear{2013}
\signatory{Dr.~Qianping Gu, Supervisor}
\signatory{Dr.~Ramesh Krishnamurti, Supervisor}

\beforepreface


\newpage


%
%

\prefacesection{Abstract}
\textsc{DORI} (Distributed Outdoor Robotic Instruments) is a remotely controlled vehicle that is designed to simulate a planetary exploration mission. \textsc{DORI} is equipped with over 20 environmental sensors and can perform basic data analysis, logging and remote upload. The individual components are distributed across a fault-tolerant bus for redundancy. A partial sensor list includes atmospheric pressure, rainfall, wind speed, GPS, gyroscopic inertia, linear acceleration, magnetic field strength, temperature, laser and ultrasonic distance sensing, as well as digital audio and video capture. The project uses recycled consumer electronics devices as a low-cost source for sensor components. This report describes the hardware design of DORI including sensor electronics, embedded firmware, and physical construction.


\lists


\beforetext

\chapter{Introduction}
Few engineering projects require as much careful planning and execution as NASA's space exploration missions. On deployment they are permanently launched to remote locations that make them completely inaccessible to further physical repairs. DORI is a simulated planetary exploration mission with self-imposed accessibility and communications constraints, in order to examine some of the engineering challenges that NASA's mission designers must overcome.

\section{Project Overview}
\subsection{Description}
\textsc{DORI} is a remotely controlled vehicle containing a distributed network of nodes able to remotely measure and record environmental data. The data are transmitted to a server for storage and manual analysis. To perform this analysis we have created a suite of data visualization tools called \textsc{tk} that allows us to filter, visualize, and interpret the data recorded by \textsc{DORI}. These tools present a standard interface for the generation of charts, graphs and plots, as well as the creation of simple 3D reconstructions of the sites where the data were acquired. We have also developed \textsc{gateway}, a central server to receive and archive the data uploaded by \textsc{DORI}. \textsc{gateway} can also automatically perform simple data preprocessing and normalization.

    \begin{figure}[h]
        \centering
        \includegraphics[width=0.6\textwidth]{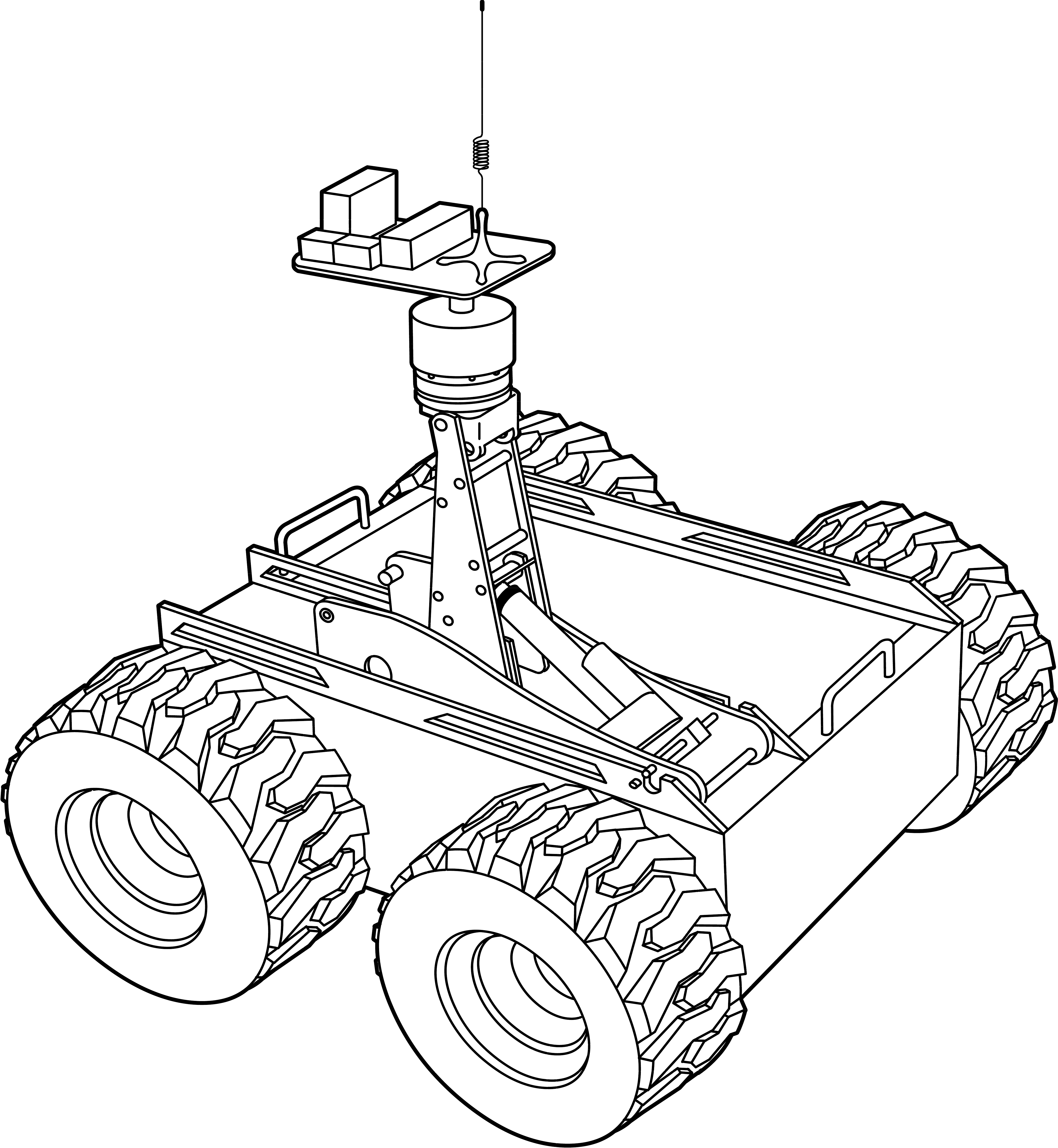}
        \caption{DORI physical design}
    \end{figure}

    DORI's main task is to collect and record the measurements output by the various nodes onto one of two SD cards (referred to as \emph{passive} operating mode). Once a large number of sensor values have been collected, one of the uplink nodes connects to \textsc{gateway} and sends the full sensor log in a single transfer. A CRC-16 checksum is then computed on both sides and these checksums are compared to confirm that the file was successfully transferred. After a successful transfer, the log file is deleted from the SD card.

    In addition to DORI's passive mode of operation there is also an \emph{active} operating mode, in which raw system bus traffic can be sent and monitored directly using either of the two uplink modems.

\subsection{Objectives}
Our goal was to create a fault tolerant distributed data collection system which can reliably operate from a location that is not easily accessible after launch. DORI will be permanently deployed to a location several hundred kilometers away to ensure we are not able to perform any physical repair after launch. In this way we hope to artificially recreate the risk and uncertainty of a space mission. We expect \textsc{DORI} to remain operational for a minimum of 12 months after deployment. After launch, we plan to conduct simple experiments using all the included sensing equipment, and to record several environmental parameters such as temperature, humidity, and precipitation.

During construction we focused on finding the cheapest possible sources for \textsc{DORI}'s hardware components. The majority of the electronic components in \textsc{DORI} were purchased from retailers in China. Most of the materials used to construct the chassis, arm, and sensor plate were salvaged from a junkyard at little or no cost. 

The processing power of \textsc{DORI} is distributed among a dozen microcontroller nodes that are each responsible for specific groups of functions. Each node manages a limited number of devices. The robot is thus protected from severe malfunction in the event of the failure of any single node.

Despite being inaccessible after deployment, \textsc{DORI} will inevitably require upgrades and maintenance after deployment at the target site. To accomplish this, we have included the ability to remotely upgrade the firmware of any node in the distributed system.

\chapter{System core}
DORI's core is comprised of a collection of nodes connected to a main system bus. The primary goal was to ensure that the main architecture was highly reliable, with redundant systems included to protect against hardware failure. 

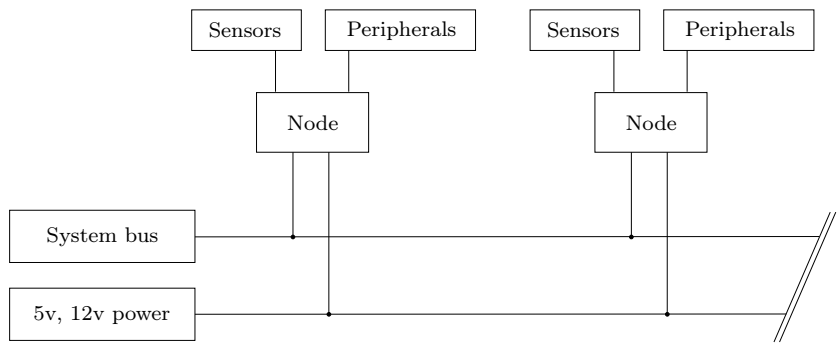
\begin{figure}[h]
    \centering
    \definecolor{cffffff}{RGB}{255,255,255}

\begin{tikzpicture}[y=0.80pt,x=0.80pt,yscale=-1, inner sep=0pt, outer sep=0pt]
\tikzstyle{every node}=[font=\scriptsize]
\begin{scope}[cm={{1.25,0.0,0.0,-1.25,(0.0,420.0)}}]
  \path[xscale=1.000,yscale=-1.000,draw=black,miter limit=4.38,line
    width=0.240pt,rounded corners=0.0000cm] (41.3070,-168.2285) rectangle
    (111.4056,-148.4549);
  \path[xscale=1.000,yscale=-1.000,fill=black] (55.290295,-154.72006) node[above
    right] (text4399) {System bus};
  \path[xscale=1.000,yscale=-1.000,draw=black,miter limit=4.38,line
    width=0.240pt,rounded corners=0.0000cm] (41.3070,-138.7317) rectangle
    (111.4056,-118.9581);
  \path[xscale=1.000,yscale=-1.000,fill=black] (50.328854,-125.29317) node[above
    right] (text4405) {5v, 12v power};
  \path[draw=black,line join=miter,line cap=butt,miter limit=4.38,line
    width=0.240pt] (111.4056,128.8580) -- (335.1013,128.9552);
  \path[draw=black,line join=miter,line cap=butt,miter limit=4.38,line
    width=0.240pt] (111.4056,158.1552) -- (347.5340,158.2832);
  \begin{scope}[cm={{1.31622,0.0,0.0,1.31682,(1.67945,-77.33039)}},miter limit=4.38,line width=0.240pt]
      \path[xscale=1.000,yscale=-1.000,draw=black,miter limit=4.38,line
        width=0.182pt,rounded corners=0.0000cm] (101.0592,-220.3038) rectangle
        (133.2734,-203.1575);
      \path[xscale=1.000,yscale=-1.000,fill=black] (109.90815,-209.69177) node[above
        right] (text4421) {Node};
      \path[draw=black,line join=miter,line cap=butt,miter limit=4.38,line
        width=0.182pt] (111.5333,203.1575) -- (111.5333,178.7671);
      \path[draw=black,line join=miter,line cap=butt,miter limit=4.38,line
        width=0.182pt] (121.8425,203.1575) -- (121.8425,156.4616);
      \begin{scope}[cm={{1.77172,0.0,0.0,1.77172,(-66.37583,-119.47174)}},miter limit=4.38,line width=0.135pt]
        \path[cm={{0.8,0.0,0.0,-0.8,(0.0,336.0)}},draw=black,fill=black,miter
          limit=4.38,line width=0.129pt] (125.8876,209.5633) .. controls
          (125.8876,209.7678) and (125.7217,209.9337) .. (125.5172,209.9337) .. controls
          (125.3126,209.9337) and (125.1468,209.7678) .. (125.1468,209.5633) .. controls
          (125.1468,209.3587) and (125.3126,209.1929) .. (125.5172,209.1929) .. controls
          (125.7217,209.1929) and (125.8876,209.3587) .. (125.8876,209.5633) -- cycle;
      \end{scope}
      \path[xscale=1.000,yscale=-1.000,draw=black,miter limit=4.38,line
        width=0.182pt,rounded corners=0.0000cm] (82.4284,-244.0851) rectangle
        (113.9330,-232.4201);
      \path[xscale=1.000,yscale=-1.000,fill=black] (87.330894,-236.26181) node[above
        right] (text4446) {Sensors};
      \path[xscale=1.000,yscale=-1.000,fill=black] (127.0157,-235.35158) node[above
        right] (text4450) {Peripherals};
      \path[xscale=1.000,yscale=-1.000,draw=black,miter limit=4.38,line
        width=0.182pt,rounded corners=0.0000cm] (120.8283,-244.0851) rectangle
        (163.9980,-232.4201);
      \path[draw=black,line join=miter,line cap=butt,miter limit=4.38,line
        width=0.182pt] (106.5148,232.5140) -- (106.5148,220.5636);
      \path[draw=black,line join=miter,line cap=butt,miter limit=4.38,line
        width=0.182pt] (127.5580,232.5140) -- (127.5580,220.5636);
      \begin{scope}[cm={{1.77172,0.0,0.0,1.77172,(-56.04266,-141.69955)}},miter limit=4.38,line width=0.135pt]
        \path[cm={{0.8,0.0,0.0,-0.8,(0.0,336.0)}},draw=black,fill=black,miter
          limit=4.38,line width=0.129pt] (125.8876,209.5633) .. controls
          (125.8876,209.7678) and (125.7217,209.9337) .. (125.5172,209.9337) .. controls
          (125.3126,209.9337) and (125.1468,209.7678) .. (125.1468,209.5633) .. controls
          (125.1468,209.3587) and (125.3126,209.1929) .. (125.5172,209.1929) .. controls
          (125.7217,209.1929) and (125.8876,209.3587) .. (125.8876,209.5633) -- cycle;
      \end{scope}
    \begin{scope}[shift={(97.24784,0)}]
      \path[xscale=1.000,yscale=-1.000,draw=black,miter limit=4.38,line
        width=0.182pt,rounded corners=0.0000cm] (101.0592,-220.3038) rectangle
        (133.2734,-203.1575);
      \path[xscale=1.000,yscale=-1.000,fill=black] (109.90815,-209.69177) node[above
        right] (text4845) {Node};
      \path[draw=black,line join=miter,line cap=butt,miter limit=4.38,line
        width=0.182pt] (111.5333,203.1575) -- (111.5333,178.7671);
      \path[draw=black,line join=miter,line cap=butt,miter limit=4.38,line
        width=0.182pt] (121.8425,203.1575) -- (121.8425,156.4616);
      \begin{scope}[cm={{1.77172,0.0,0.0,1.77172,(-66.37583,-119.47174)}},miter limit=4.38,line width=0.135pt]
        \path[cm={{0.8,0.0,0.0,-0.8,(0.0,336.0)}},draw=black,fill=black,miter
          limit=4.38,line width=0.129pt] (125.8876,209.5633) .. controls
          (125.8876,209.7678) and (125.7217,209.9337) .. (125.5172,209.9337) .. controls
          (125.3126,209.9337) and (125.1468,209.7678) .. (125.1468,209.5633) .. controls
          (125.1468,209.3587) and (125.3126,209.1929) .. (125.5172,209.1929) .. controls
          (125.7217,209.1929) and (125.8876,209.3587) .. (125.8876,209.5633) -- cycle;
      \end{scope}
      \path[xscale=1.000,yscale=-1.000,draw=black,miter limit=4.38,line
        width=0.182pt,rounded corners=0.0000cm] (82.4284,-244.0851) rectangle
        (113.9330,-232.4201);
      \path[xscale=1.000,yscale=-1.000,fill=black] (87.330894,-236.26181) node[above
        right] (text4859) {Sensors};
      \path[xscale=1.000,yscale=-1.000,fill=black] (127.0157,-235.35158) node[above
        right] (text4863) {Peripherals};
      \path[xscale=1.000,yscale=-1.000,draw=black,miter limit=4.38,line
        width=0.182pt,rounded corners=0.0000cm] (120.8283,-244.0851) rectangle
        (163.9980,-232.4201);
      \path[draw=black,line join=miter,line cap=butt,miter limit=4.38,line
        width=0.182pt] (106.5148,232.5140) -- (106.5148,220.5636);
      \path[draw=black,line join=miter,line cap=butt,miter limit=4.38,line
        width=0.182pt] (127.5580,232.5140) -- (127.5580,220.5636);
      \begin{scope}[cm={{1.77172,0.0,0.0,1.77172,(-56.04266,-141.69955)}},miter limit=4.38,line width=0.135pt]
        \path[cm={{0.8,0.0,0.0,-0.8,(0.0,336.0)}},draw=black,fill=black,miter
          limit=4.38,line width=0.129pt] (125.8876,209.5633) .. controls
          (125.8876,209.7678) and (125.7217,209.9337) .. (125.5172,209.9337) .. controls
          (125.3126,209.9337) and (125.1468,209.7678) .. (125.1468,209.5633) .. controls
          (125.1468,209.3587) and (125.3126,209.1929) .. (125.5172,209.1929) .. controls
          (125.7217,209.1929) and (125.8876,209.3587) .. (125.8876,209.5633) -- cycle;
      \end{scope}
    \end{scope}
  \end{scope}
  \path[draw=black,line join=miter,line cap=butt,miter limit=4.38,line
    width=0.240pt] (351.7073,167.5268) -- (330.4212,118.3501);
  \path[draw=black,line join=miter,line cap=butt,miter limit=4.38,line
    width=0.240pt] (353.8132,167.5268) -- (332.5272,118.3502);
\end{scope}
\end{tikzpicture}
    \caption{Core system design}
\end{figure}

    \section{Message bus}
    DORI's main system backbone is a Controller Area Network (CAN) bus with a frequency of 125 kbit/s. Short messages are broadcast between all nodes, and bus arbitration is achieved via a priority system which ensures that no collisions occur and no packets are lost. This also guarantees message delivery subject to certain timing constraints~\cite{cantimes}.

    The CAN bus protocol (ISO 11898) was developed in the 1980s at Robert Bosch GmbH for various automotive and industrial automation applications. It is a robust, multi-master broadcast bus that uses differential NRZ (non-return-to-zero) signallining on two wires. CAN messages consist of an 11 or 29-bit long message identifier and up to 8 bytes of data payload. 

    We began by evaluating existing higher-level protocols such as \emph{J1939} and \emph{CANopen}, which automatically fragment messages longer than 8 bytes across multiple CAN packets. We found that these are highly complex, closed standards that provide more features than necessary for DORI's simple distributed design. We eventually decided to use a packet format similar to J1939's multipacket \emph{transport protocol} specification, with custom modifications to allow for very large data transfers such as JPEG images and firmware updates.

    Atmel produces several AVR microcontrollers built specifically for automotive applications with integrated CAN controller logic. The primary disadvantage with these products is that they are only offered in small surface-mount packages (TQFP and VQFN) which are difficult to use in breadboard-based development. These chips also tend to be more difficult to purchase in small quantities, and cannot be purchased from local suppliers due to small hobbyist demand. Note that these Automotive AVR devices do not feature integrated CAN transceivers, and these must still be purchased separately.

    \section{Uplink}
    The uplink connection used by DORI to communicate with our \textsc{gateway} server must always be available to receive new commands and transmit sensor readings. If the uplink were to fail we would neither be able to read the sensor values nor diagnose the failure. We therefore decided to install two separate uplink modems: one for normal use, and another for emergency communications.

    Several communications strategies were evaluated for DORI's main uplink. We first decided to use an RF transceiver to send and recieve Olivia MFSK-encoded AX.25 data over the HF frequency range using hobbyist HAM equipment~\cite{olivia-mfsk}. This packet radio format would give us 1200 baud bidirectional transfer at distances of hundreds of kilometers using NVIS (skywave) propagation. We also attempted to design a mechanism where DORI would deploy a 7.5m monopole antenna directly onto the ground to transmit on the 5MHz/60m band. Eventually we decided that this solution had too many drawbacks including weather interference, low transmission range, and high power usage, and it was abandoned.

    Because the communication link was critical in this project, we decided to piggyback on the existing cellular mobile network infrastructure that is installed throughout the country. Two GSM cellular modems were purchased: a \brand{BenQ M32} USB GSM modem as the main modem, and a \brand{Nokia 3390} cellular phone with SMS capability as a backup modem.
    
    \begin{wrapfigure}{l}{0.17\textwidth}
        \centering
        \includegraphics[height=1cm,angle=270,origin=c]{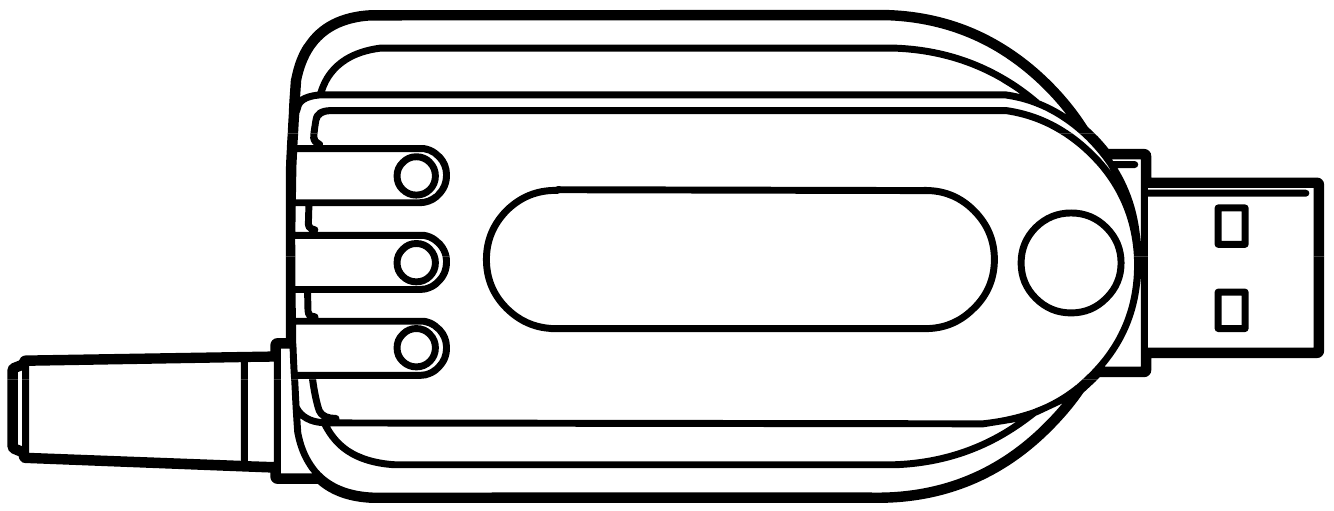}
        \caption*{BenQ M32}
    \end{wrapfigure}

    DORI communicates with our \textsc{gateway} server on the cellular network using its main modem, which uses the \brand{7-11 SpeakOut} network at 850/1900MHz. The module's existing antenna was a simple stub of metal, so it was replaced with an external GSM antenna meant for vehicle use, increasing its signal strength by approximately 2dB. The module includes a \brand{Prolific PL2303} USB-serial controller which we bypassed in order to directly access the \brand{M32} chip's 3.3v UART input and output signals~\cite{m32}.

    The \brand{SpeakOut} network offers limited web browsing using an HTTP proxy. It does not support data (raw TCP) connections, WebSockets, or any form of HTTP long-polling. However we found that DNS lookups are not verified, and TCP port 53 is open and available for external connections. Using the service's \$10/month ``unlimited browsing'' package, we are able to communicate with and transfer data from DORI at a fixed low monthly cost.

    \begin{wrapfigure}{l}{0.17\textwidth}
        \centering
        \includegraphics[height=2cm]{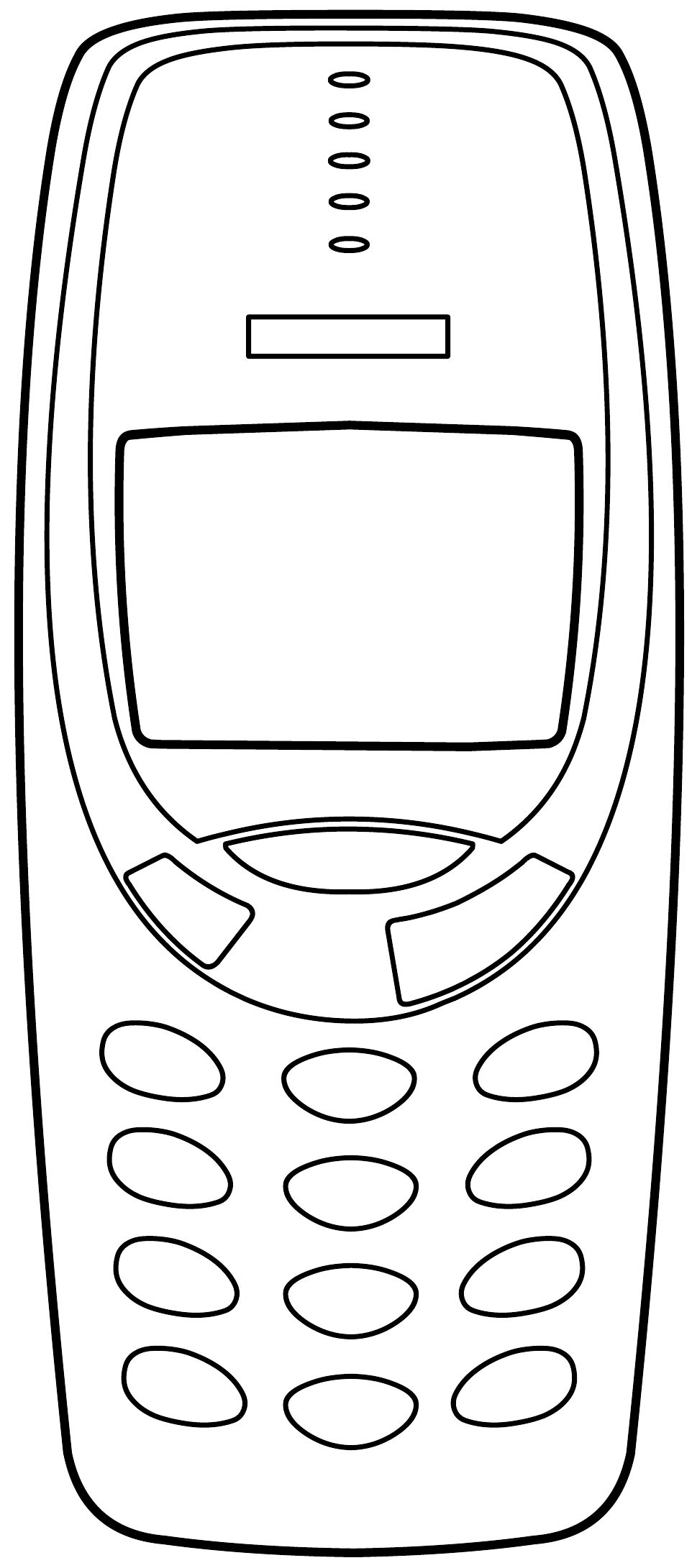}
        \caption*{Nokia 3310}
    \end{wrapfigure}
    
    As a backup communication method, DORI is also equipped with a Nokia 3390 cellular phone with SMS capability using a \brand{Fido Mobility} SIM card. The secondary modem node connects directly to the phone's \brand{FBus} (Fast Bus) serial communication port, and is used as a simple bridge interface to allow us to send and receive raw packets on DORI's system bus. In the event of a communication failure on the main modem, the operator will begin by sending a broadcast system message to the secondary modem node to temporarily disallow all periodic sensor broadcasts, followed by a command to begin forwarding all system bus traffic over the secondary node's SMS interface. In this way we are able to perform any necessary operations including analyzing the state of any node in the system, requesting sensor values or file contents, and even perfoming emergency firmware reflashing.

    \section{Nodes}
        \subsection{AVR}
        Each node consists of an \brand{Atmel ATmega88} 8-bit RISC microcontroller with 8192 bytes of flash program memory and 1024 bytes of RAM~\cite{atmega88} connected to a number of external sensors and peripherals. Every node runs a custom event loop configured to the specific peripherals controlled by it rather than a generic task-switching operating system. Due to the lightweight nature of the driver software and event loop system, most of our node firmware is smaller than 2000 bytes (notable exceptions are the FAT16 driver and the NMEA datestamp parsing code.)

        Node firmware is deliberately simple, and very little pre-processing is performed by DORI. Most drivers are simple wrappers which read sensor values, perform simple low-pass filtering and normalization. These semi-processed sensor values are then directly broadcast on the system bus to be logged and uploaded for analysis. All CPU-intensive processing is performed by \textsc{gateway} and \textsc{tk}. We kept each node's electrical circuit as small as possible, using minimal components to interface the hardware's signal with its node's microcontroller.
        
        Nodes are powered by a dual 12v/5v regulated supply. Any node which must interface with peripherals running at voltages other than 5v or 12v has its own power regulator based on the LM2596 step-down (buck) regulator. In particular, nodes interfacing with a large number of 3.3v peripherals have their AVR microcontroller powered by 3.3v, and only the \brand{MCP2515/MCP2551} CAN interface circuitry is powered by 5v. A zener diode clipper circuit is then used to interface the \brand{MCP2515}'s 5v signals with the AVR's 3.3v inputs.

        \subsection{Data reporting}
        As most of DORI's sensor hardware is built from inexpensive consumer-grade equipment, the resulting sensor values are relatively noisy and should be filtered before they are transmitted back for further processing and analysis. Several filter algorithms were considered with respect to algorithm complexity, smoothing performance, and compiled code size. A common filter technique is known as a \emph{linear rolling-average} filter, where the most recent $n$ samples are averaged and the resulting mean values of these samples is a low-pass filtered copy of the input signal ($\mathit{out}_i = \mathit{in}_{i-n} + \ldots + \mathit{in}_{i}$.) A variant of this technique, known as \emph{exponential rolling average}, applies new data values to a rolling average value which undergoes exponential decay ($\mathit{out}_i = \alpha \mathit{in}_i + (1 - \alpha) \mathit{out}_{i-1}, \,0 \le \alpha \le 1$). In DORI we use a 1-dimensional Kalman filter, which is a linear filter that can be precisely tuned to minimize a particular noise distribution~\cite{kalman}.

\chapter{Construction}
    DORI was designed as an experimental low-cost robot architecture. Our focus was to develop a highly reliable distributed sensor network, using limited hardware resources. As such, DORI's chassis, drive system, and actuator arm were kept minimal and simplistic.
    \section{Chassis}
The main chassis was constructed primarily of salvaged materials. An old streetlight enclosure was used for the frame, a broken floor jack was used as the actuator arm, and the wheels were taken from an old farm tractor. The two chassis components that could not be sourced from a junkyard were the linear actuator used to raise and lower the arm and the stepper motor to rotate the sensor plate. These were bought online from China at low cost.

    \section{Drive system}
    DORI uses a simple drive system consisting of two electric motors driving four wheels. The motors are 12v power window motors from a vehicle, with built-in 20:1 gear reduction. The motors are directly connected to the drive shafts, which are lawn mower drive axles giving a further 6:1 gear reduction, for a total gear ratio of 120:1. The wheels on each side of the robot are directly linked together with a chain giving us simulated four-wheel drive using only two motors. This drive configuration is similar to a tank, and having a zero turning radius greatly simplifies our ability to control DORI's position. Large rubber tractor wheels are used in order to navigate the rough terrain of the target environment.

    \section{Actuator arm}
    The actuator arm is built from a car floor jack that is raised and lowered by a 12v linear actuator. Our priority was to make sure that the arm was physically secure and would not be affected by debris or strong winds. The linear actuator has a rated dynamic lifting capacity of 50kg, but this capacity is reduced due to the linear actuator's cantilever configuration with the arm beam. During testing we were able to raise a 40kg bag of sand with no noticeable strain on the lifting mechanism. The static holding strength of the linear actuator is 226kg when power is removed, so the arm is unlikely to drift from a raised position when supporting the sensor plate and attached instruments (5kg). During testing we lowered the arm to a near-horizontal position and placed a 40kg bag of sand on top for 12 hours, with no measurable change in height.

    The linear actuator we purchased has an integrated precision linear potentiometer which is able to measure the position of the actuator arm. We use this as a feedback mechanism to report the arm's position while it is raised or lowered. The arm's position is also periodically monitored while the arm is static to immediately notify \textsc{tk} of any gradual drift that may occur.

    \section{Power}
    DORI is powered by two 12v vehicle batteries each rated to deliver 54Ah. One battery is used to power the electronic components, and the other battery is used to power all of the high-current power electronics. In this configuration they are completely isolated from each other, and the sensors are completely protected from electrical noise from the motors. The batteries are constantly recharged by a 10W solar panel during daylight hours. The electronic circuits draw little current -- the main system bus and the nodes themselves could run for years on a single charge. The biggest sources of power draw are the drive motors and linear actuator, which draw up to 3A when stalled. However DORI has no self-navigation abilities and must be manually controlled. The surrounding environment must be studied using \textsc{tk} and a path chosen before any drive commands are sent to DORI. Therefore DORI will usually stay in a single spot and record measurements. The motors will be rarely used so the power they draw will be minimal over time.

    \section{Nodes}
    Every node is constructed from an identical base circuit layout, with specific hardware modifications added to accomodate the particular peripherals connected to a particular node. The main processor in each node is an Atmel ATmega88 chip clocked from an 8MHz quartz crystal oscillator. Each node connects to the system bus using a Microchip MCP2515 CAN controller connected to a MCP2551 CAN transceiver.

    The nodes are distributed throughout DORI according to the specific peripherals they control. A piece of stranded 10AWG (5 mm$^2$) wire is used to transmit system bus messages across the nodes. Each node uses a pair of wire tap connectors to physically connect to the system bus, and these electrical taps are wrapped in insulating electrical tape to prevent corrosion.

    \section{Durability}
We intend to operate \textsc{DORI} for a minimum of 12 months after deployment, so physical reliability is critical. \textsc{DORI} will be operating in a densely forested environment in temperatures ranging from $-6\degrees{}$ to $24\degrees{}$, with an average snowfall of over 400cm.

Other possible hazards include rats, water condensation, and electrical interference due to thunderstorms. Rats love to chew wires, so we have used steel armored cable wherever possible for protection. Most of the signals within \textsc{DORI} are relatively low frequency, so they are less susceptible to EM interference.

In order to avoid any moisture condensation inside the electronic components, each moisture-sensitive device underwent a waterproofing procedure. We first placed dessicant packets from ramen noodles into any large enclosed spaces, and then used a home hair dryer to heat and dry the air that would remain sealed within the electronics. Once the air was sufficiently dry we used a rubberized chemical spray to completely coat the electrical components in a water-tight rubber barrier.

\chapter{Peripherals}
    DORI is equipped with several types of peripherals for sensing external conditions, logging data, and controlling output devices such as motors and relays. We have also included peripherals which can be used to help assess and maintain DORI's physical condition.

\section{Sensors}
    DORI's primary function is to log atmospheric, meteorological, and other environmental data through various sensors. These sensor devices were obtained from various sources, including hobbyist electronic modules, broken consumer electronics, and handheld measurement devices. They are distributed among DORI's processing nodes according to their function, as well as their installed location within DORI's chassis.
    
    A digital screen reading technique was developed in order to extract digital sensor readings from handheld consumer sensing equipment such as distance meters and infrared thermometers. We remove the equipment's Liquid Crystal Display (LCD), and the digital signals output by the equipment's main circuit board are connected to a microcontroller. We can then analyze the signal being fed to the LCD, and reinterpret the digits that were being displayed for a particular measurement. This allows us to acquire calibrated digital measurements directly off of various sensor devices, which simplifies the task of interfacing with  different types of analog sensor elements. We take advantage of the existing engineering and design solutions within these devices to reduce the effort and cost associated with implementing the various sensors used within DORI. Some devices use low-cost \emph{static LCD} technology with negative voltages used to control the display segments. For these devices we use a silicon-germanium diode clamping circuit to limit these voltages to -0.3v in order to protect the microcontroller's inputs.

\subsection{Atmospheric data}
The primary mission objective is to record remote environmental measurements and transfer the data back to \textsc{tk}. The atmospheric sensors are therefore very critical in studying the environment of the target site. The atmospheric sensor nodes are configured to output continuous periodic measurements, and the output rate can be changed to affect the granularity of the reported measurements.

    \subsubsection*{Temperature}
    \begin{wrapfigure}{l}{0.17\textwidth}
        \centering
        \includegraphics[height=1.8cm]{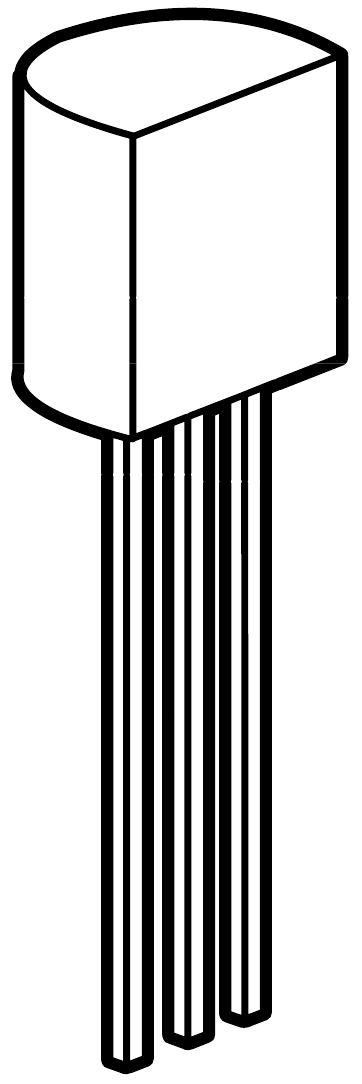}
        \caption*{DS18B20 sensor}
    \end{wrapfigure}
    We decided to use the \brand{Maxim DS18B20} 1-Wire Digital Temperature Sensor due to our prior experience with this particular device. This sensor has a measurement range of $-55\degrees{}$ to $+125\degrees{}$, with $\pm0.5\degrees{}$ accuracy from $-10\degrees{}$ to $+85\degrees{}$~\cite{ds18b20}. It is pre-calibrated and can be connected using a single wire simultaneously carrying data and power. Several DS18B20 sensors were acquired as free samples from Maxim Semiconductor, and are installed among the various components to gather various temperature readings.

    Maxim's 1-Wire devices can be individually addressed over a shared data wire, referred to as a 1-Wire network. The 1-Wire search algorithm is conceptually similar to the bus arbitration used by CAN. Individual bits are sequentially queried across each possible device identifier, and any matching devices are further specified until they can be uniquely identified by the bits in the query~\cite{1wiresearch}. Using this search algorithm, we can place hundreds of 1-Wire sensors across the robot, all interfaced via a single system bus node. For our purposes we decided to include 12 temperature sensors placed inside and outside the robot frame, as well as scattered among the heat-sensitive electronic components, and one temperature sensor for each of the four motors (stepper motor, linear actuator, and both drive motors.)

    \subsubsection*{Air pressure}
    \begin{wrapfigure}{l}{0.2\textwidth}
        \centering
        \includegraphics[height=1.9cm]{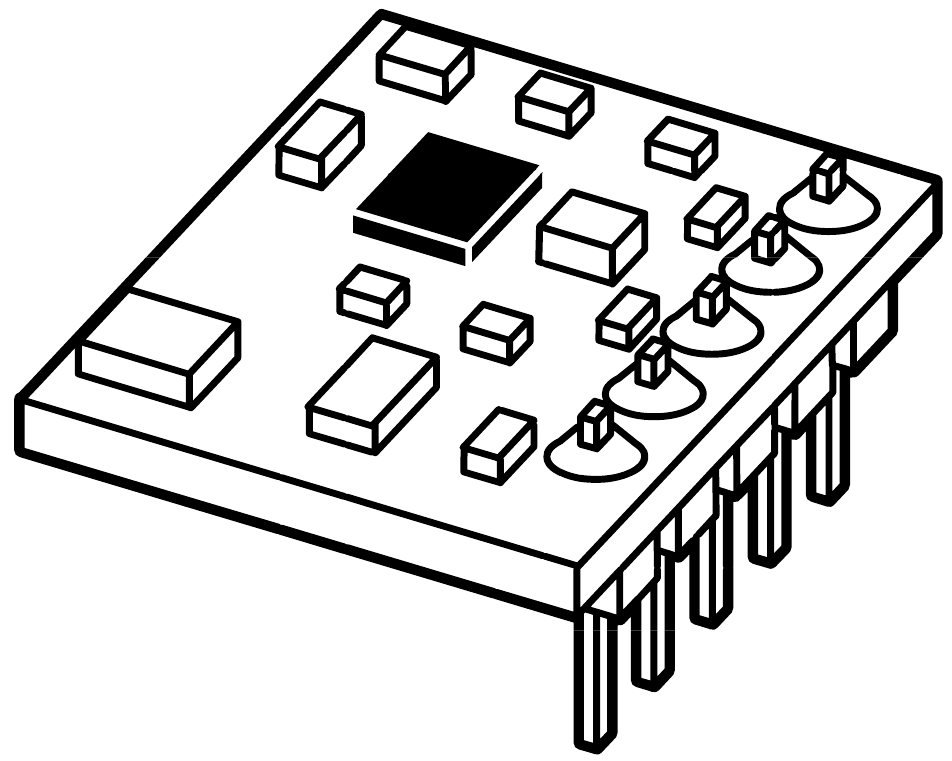}
        \caption*{BMP085 module}
    \end{wrapfigure}
    We picked the \brand{Bosch Sensortec BMP085} barometric pressure sensor because of the many inexpensive hobbyist-level development modules for the sensor available from online suppliers. The BMP085 is an \textsc{i2c} sensor which outputs a raw pressure value from a piezo-resistive sensing element. A compensation algorithm is also provided to convert the raw sensor values to calibrated barometric pressure measurements~\cite{bmp085}. The output from this sensor represents the current air pressure in kPa, and two additional measurements can be derived from it. When this value is averaged over several days it can be used as a high-resolution altimeter. In addition, the current barometric pressure reading can then be compared to the average pressure for a particular altitude to provide a rough estimate of changing weather conditions such as incoming storms or other pressure systems.

    \subsubsection*{Wind velocity}
There are many common wind speed and direction sensor designs, most of which use a rotating wheel anemometer to measure wind speed combined with a wind vane to measure wind direction. These designs are delicate and difficult to seal against moisture and debris. There are also cheap handheld digital units such as the \textsc{GM8908} for \$14, but most of these are unidirectional~\cite{gm8908}.

We adopted a basic solid-state design in order to simplify the hardware configuration. We attached an aftermarket Wii Nunchuck containing an Analog Devices \textsc{ADXL335} accelerometer to the end of a flexible plastic tube mounted vertically to the robot's body. The tube is further wrapped in thick foam to increase its wind resistance.

\begin{figure}[h]
    \centering
    \scriptsize
    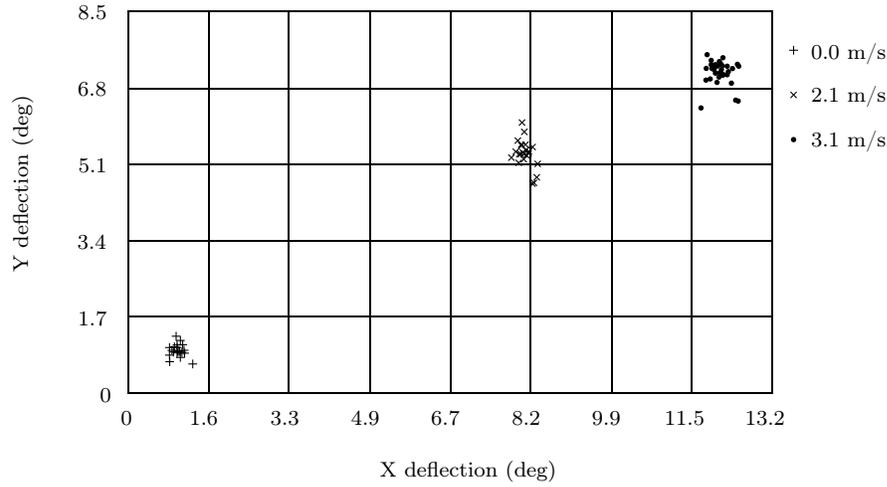
    \caption{Wind velocity sensor calibration data}
\end{figure}

The device can be modelled as a beam with a uniformly distributed load: as the tube gets pushed by the wind, the deflection is measured by the accelerometer mounted at the top. Using this sensor construction means that the entire device is environmentally sealed from moisture, dirt, and other debris. However this simple design means the sensor's output is highly nonlinear and must be calibrated against several known values before useful wind speed measurements can be recorded. Therefore, we have taken several sample readings using a large electric fan, and calibrated them against simultaneous values from a \textsc{GM8908} unit~\cite{gm8908}. We must also compensate for the effect of gravity by subtracting the current tilt of the robot from the wind sensor accelerometer's values.

    \subsubsection*{Rainfall}
    \begin{wrapfigure}{l}{0.3\textwidth}
        \centering
        \includegraphics[height=2cm]{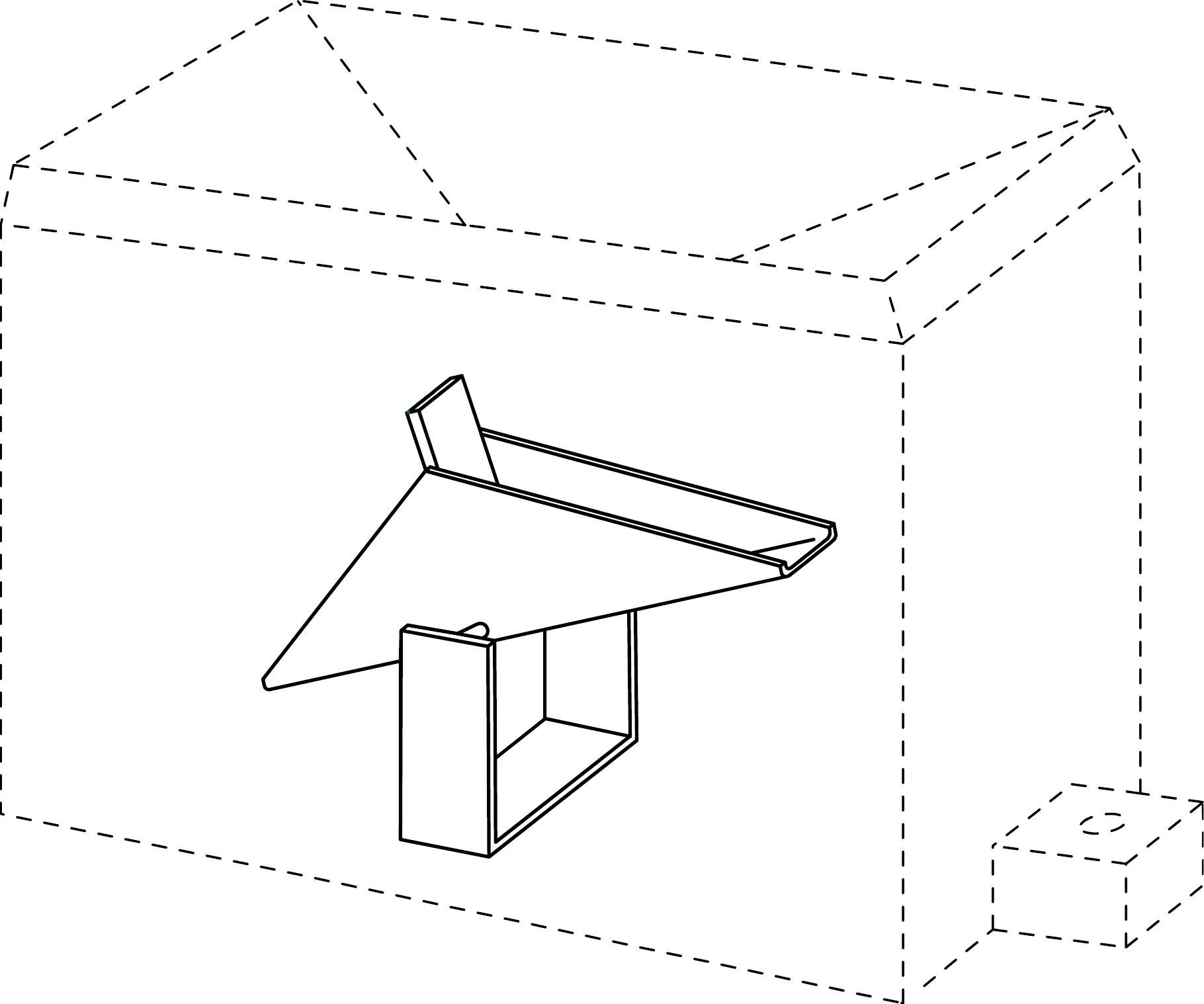}
        \caption*{WH-0531 rain gauge}
    \end{wrapfigure}
The target exploration site is a coastal rainforest, so rainfall will be a very important environmental measurement. Like the wind velocity sensor, there are several common designs used by hobbyists to measure rainfall. The simplest design measures the water height in an open-ended container. This design is prone to failure if heavy rainfall fills the container faster than it is allowed to drain, causing it to overflow. Another common design uses a \textit{tipping cup} design, where rain is funneled into one of two cups mounted on opposite ends of a tipping lever. Once a cup is filled, the weight of the water inside causes the cup to tip down and spill its contents, raising the other cup to be filled - a small sensor mounted on the lever arm records each tip event as a fixed volume of rain that has fallen since the device last tipped. However this design also has flaws, most notably a lack of precision as well as increased mechanical complexity.

Due to the target site's high rainfall (an annual average of 2400 mm~\cite{rainfall}), we decided to use the tipping cup design. We found a consumer-level device, the \textsc{WH-0531} Wireless Rain Gauge, which uses a simple plastic tipping cup sensor. To measure frozen precipitation such as snow and hail, we modified the rain gauge by adding a heating pad to the underside of the collecting funnel. This heating pad will be activated whenever the external temperature falls below $0\degrees{}$, melting the snow so that it can be measured.

    \subsubsection*{Rain pH}
DORI has two separate pH measurement strategies. The rain captured by the rain gauge flows past a saturated calomel (mercury (I) chloride) sensing element. In addition, a strip of litmus paper is attached to DORI's frame, exposed to any falling rain. After DORI is deployed to the target site, we will monitor the rainfall sensor until we detect rain, and then photograph the litmus paper using the digital camera.

We use the \brand{SKU012120} Digital pH Sensor which can measure pH values between 0 - 14 with an accuracy of approximately $\pm0.2$ pH. This is a consumer product meant for household pH measurement of pools and aquariums which we have found to be very reliable and accurate.

The second pH measurement device we include is a strip of universal indicator litmus paper that can measure pH values between 1 - 14. We have also mounted a color swatch beside the litmus paper for colour calibration. This can be used to calibrate the color balance of the photos taken by the camera to get a more accurate measurement of the color of the litmus paper once it has been exposed to rainwater. The paper can only be used once, but this will allow us to get a single reference pH measurement of the rainwater in the target environment. 

    \subsubsection*{Smoke}
    \begin{wrapfigure}{l}{0.17\textwidth}
        \centering
        \includegraphics[height=2.2cm]{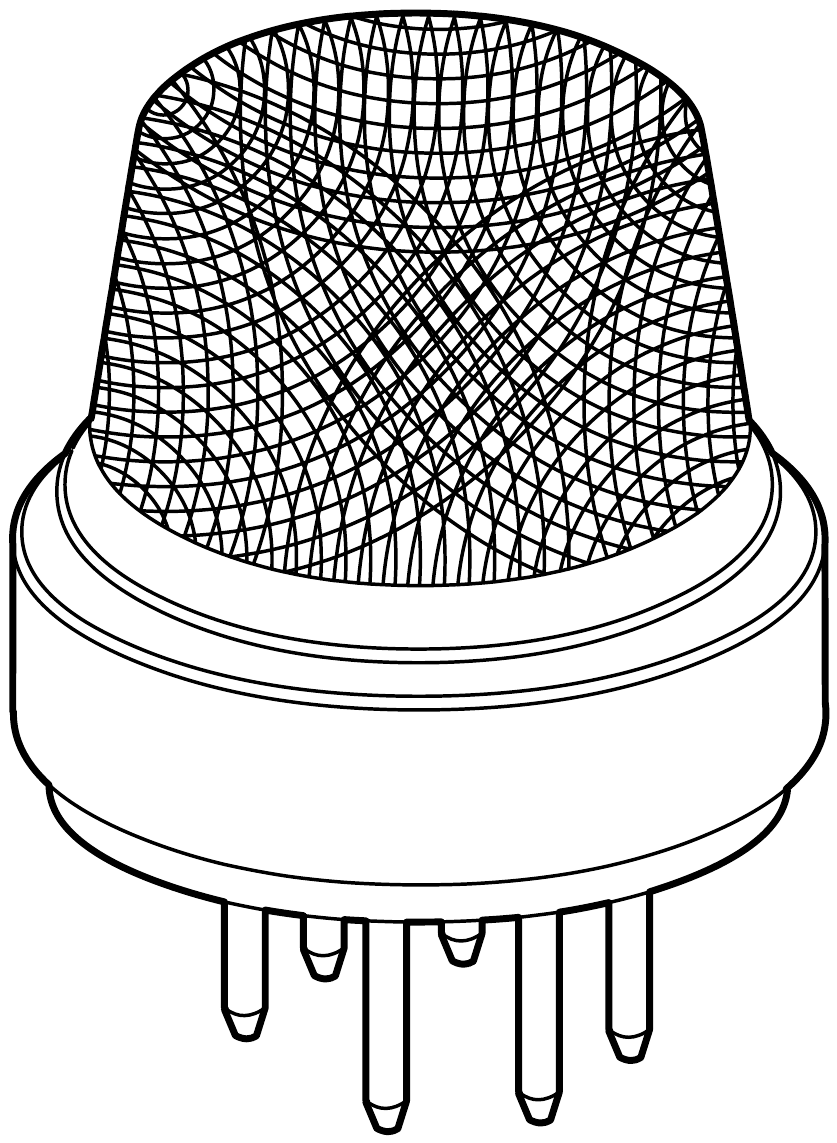}
        \caption*{MQ-2 sensor}
    \end{wrapfigure}
    We have also included a \brand{MQ-2} combustible gas sensor, which is sensitive to methane, hydrogen, liquified petroleum gas such as propane and methane, and smoke particulates. We do not expect to find many flammable gases in the air at the target site, but it is possible that a nearby forest fire could be detected based on elevated smoke particulate measurements. However we lack the necessary equipment to calibrate this sensor's analog output, so it is useful only as a relative measurement. If we graph this value over time, we may be able to detect trends which will help us detect unexpected changes in the composition of the air.

    \subsubsection*{Humidity}
    \begin{wrapfigure}{l}{0.16\textwidth}
        \centering
        \includegraphics[height=2.7cm]{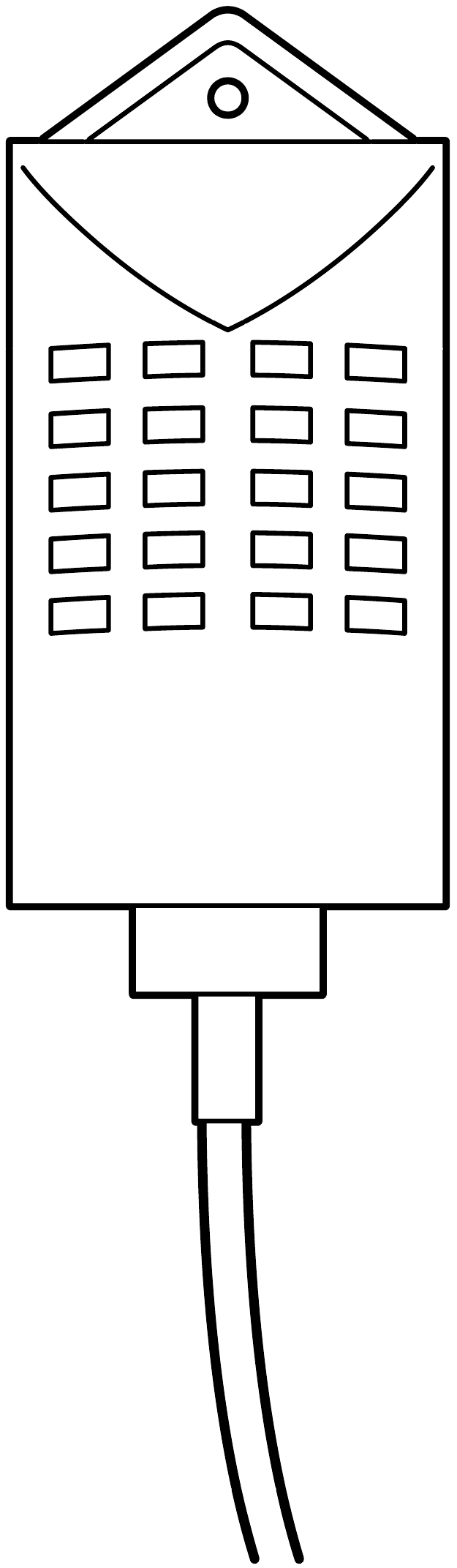}
        \caption*{CHM-02 sensor}
    \end{wrapfigure}
    We purchased a \brand{CHM-02-L} dual temperature and humidity sensor from an electronics market in China. Later we realized that the \brand{LM35} temperature sensor had not been included on the board so this sensor is only usable as a humidity sensor (this configuration should be sold as \brand{CHM-02}.) The sensor element is a \brand{Cybersen CHR02-233} based on an Al$_2$O$_3$ substrate sensing element, connected to a \brand{TI LM2902} op-amp. As configured, the sensor outputs a voltage ranging from 0.3 - 2.7v linearly corresponding to humidity values of 10\% - 95\%. 

\begin{figure}[h]
    \scriptsize
    \centering
    \includegraphics{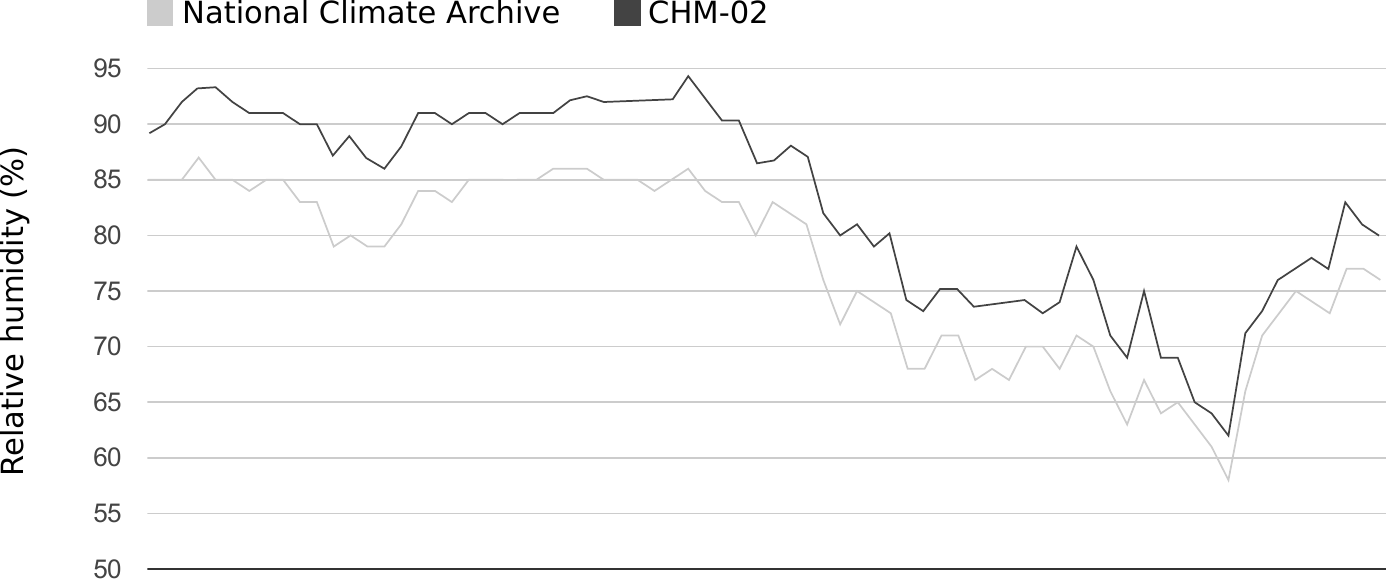}
    \caption{Calibration of CHM-02 against Environment Canada climate data~\cite{humidity}}
\end{figure}

\subsection{Telemetric data}
    DORI is also equipped to measure geomatic data about its environment. It is able to record distance and location information to allow the operator to perform basic surveying and cartography using \textsc{TK}.

\newpage
    \subsubsection*{Laser distance}
    \begin{wrapfigure}{l}{0.2\textwidth}
        \centering
        \includegraphics[height=4.5cm]{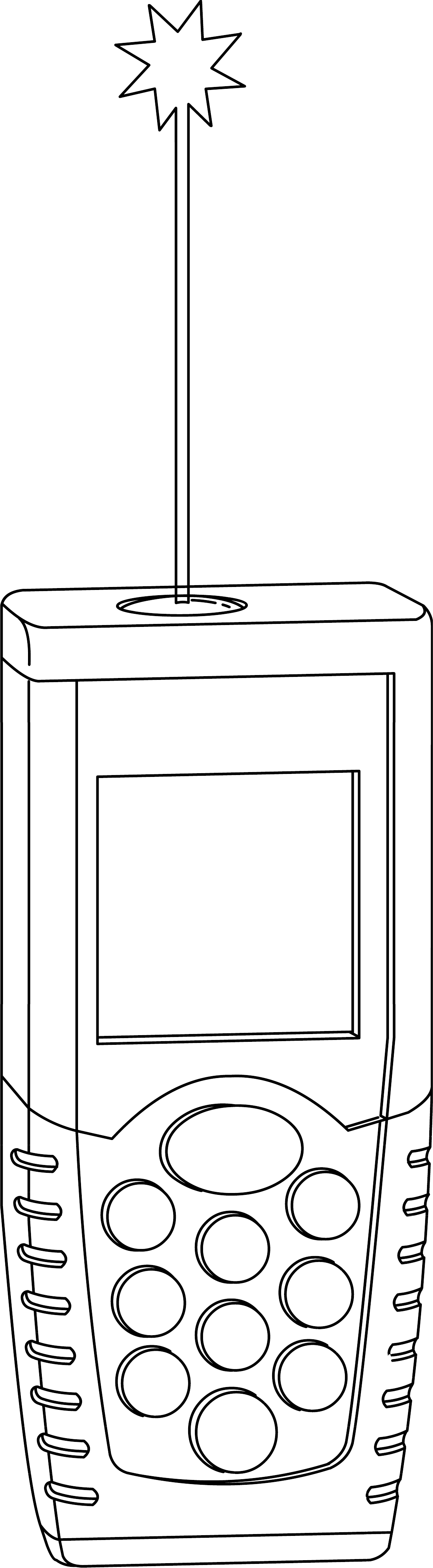}
        \caption*{LDM-40 unit}
    \end{wrapfigure}
    The general name used for robot distance measurement technologies is called LIDAR (Light Detection and Ranging). Common LIDAR techniques include using a combination of lasers, mirrors, and other optical scanners to measure the length of time taken for a beam of light to reflect off a distant object. This technique requires extremely high precision electronic circuits as well as highly calibrated optical sensing equipment.

    An example of consumer-grade laser distance scanning is the \brand{Neato Robotics XV-11} robotic vacuum where LIDAR is used for indoor navigation and mapping. A laser is aimed at a spinning mirror able to direct the beam in a full $360^\circ$ circle around the robot. Distance measurements are continually taken and a two-dimensional point cloud is created represented a single horizontal plane of the environment. Shortly after its release, the \brand{XV-11}'s sensor module was reverse-engineered and information was published allowing hobbyist electronics enthusiasts to reuse the \brand{XV-11}'s LIDAR sensor in their own projects. The biggest drawback of this technique is its cost, as the \brand{XV-11} is sold for over \$300 and \brand{Neato} has not announced plans to sell the sensor module separately.

    Another LIDAR technique that is used by the \brand{Microsoft Kinect} sensor is to project a pattern of infrared points onto a scene which are viewed by one or more cameras. The position of the infrared points can be measured to reconstruct the depth information of the original scene. However this system produces large amounts of data and requires a fast processor to perform the three-dimensional reconstruction. The Kinect sensor itself has a USB 2.0 High-speed interface and produces too much data for an 8-bit microcontroller to process.

    In order to avoid the engineering challenges of creating a precision LIDAR system, we decided to use a handheld laser distance meter as our sensor element. We purchased a \brand{CEM LDM-40} Laser Distance Meter from a Chinese supplier, capable of measuring distances from 0.2 - 40 meters with a resolution of approximately 2 mm. This handheld unit is designed to display the measured distance on an LCD. We inspected the signals being sent to the LCD and discovered that the individual display segments are controlled using SPI. We analyzed the protocol and determined how each display segment is controlled. In this way, we were able to capture the number being displayed on the screen and use it for telemetric measurements.

    \subsubsection*{Ultrasonic distance}
    \begin{wrapfigure}{l}{0.28\textwidth}
        \centering
        \includegraphics[height=2.2cm]{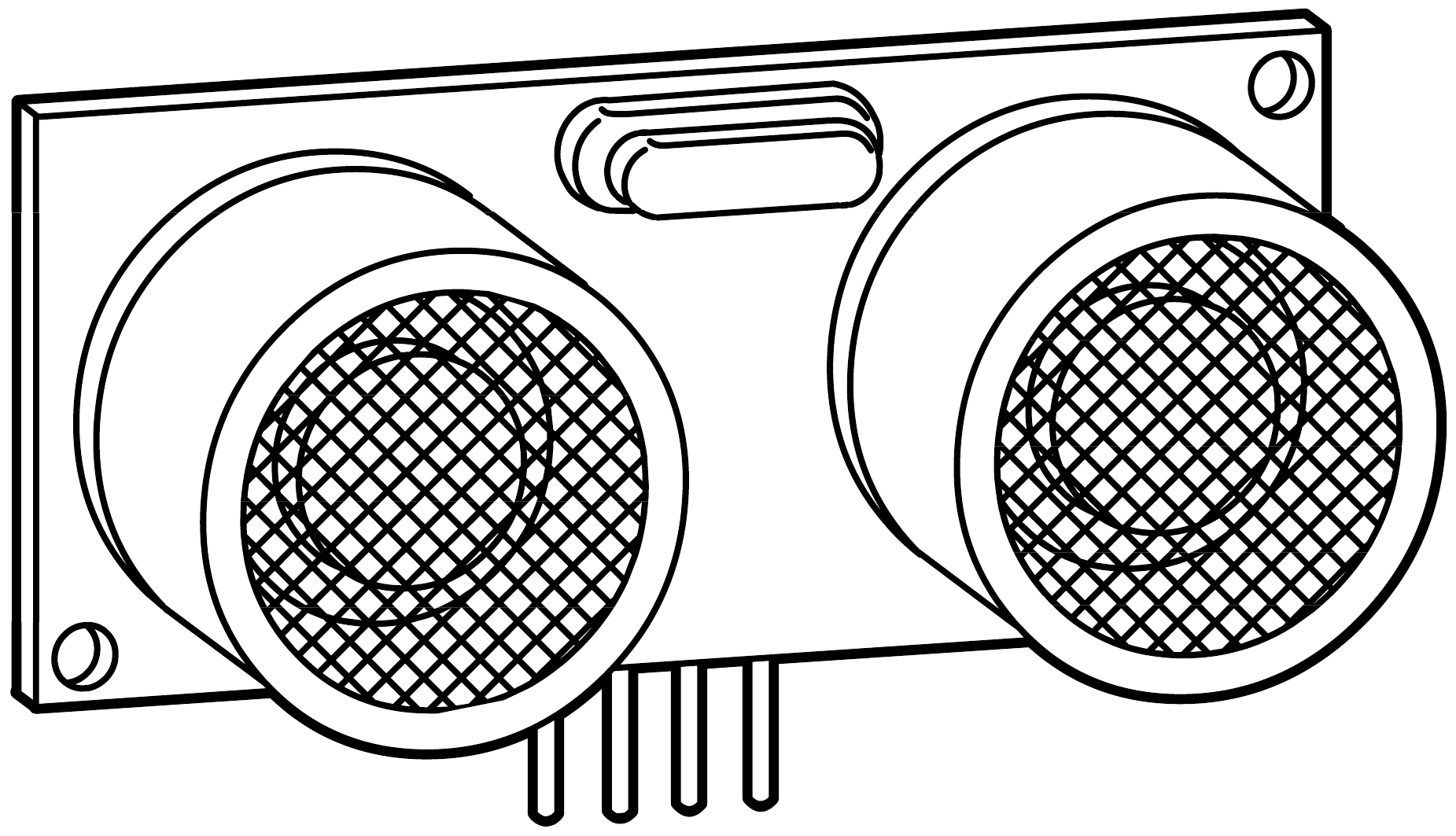}
        \caption*{SC04 module}
    \end{wrapfigure}
    The cheapest way to acquire ultrasonic distance range modules is from online suppliers, and there are several ultrasonic modules which share a compatible three- or four-pin interface such as the \brand{SR04}, \brand{SRF05}, \brand{DYP-ME007}, and the \brand{Parallax PING)))} sensor. The sensing technique used by all of these units is the same: a high-frequency (ultrasonic) audio pulse is emitted from one transducer head, and any returning echo pulse is received and filtered by a second head. The output signal's logic high state drops to 0v once the echo pulse is detected, and the length of this high state can be divided by the approximate speed of sound in air at a typical temperature (343 m/s for $20\degrees{}$) to obtain a distance estimate ranging from 5cm up to several meters depending on the reflectivity of the material. This distance measurement represents the distance travelled by the sound pulse, and can be divided by 2 to give the distance to the object. If higher precision is needed then the current environmental temperature can be taken into account and an updated value can be used for the speed of sound.

    This module cannot be waterproofed by means of the same technique used on simpler devices (see "Construction - Nodes") due to the mechanical interference caused by the waterproof coating on the sound pulse emitted by the sensor. As a cheap solution we decided to use the ultrasonic sensor elements found in cars which are used to give the driver an audible proximity warning signal when they approach an obstacle while backing up their car. We visited a junkyard and took two waterproof ultrasonic transducer sensors from a wrecked car bumper we found in the garbage. We bought a standard \brand{SR04} ultrasonic module from an online retailer and replaced its indoor ultrasonic sensors with the car bumper sensors. We removed the indoor ultrasonic sensor elements from the \brand{SR04} and replaced them with the waterproof vehicle sensors. This gives us a rugged, environmentally-sealed distance measuring device at a fraction of the cost of a waterproof sensing module sold by robotic parts vendors.

    \subsubsection*{Infrared thermometer}
    \begin{wrapfigure}{l}{0.25\textwidth}
        \centering
        \includegraphics[width=0.22\textwidth]{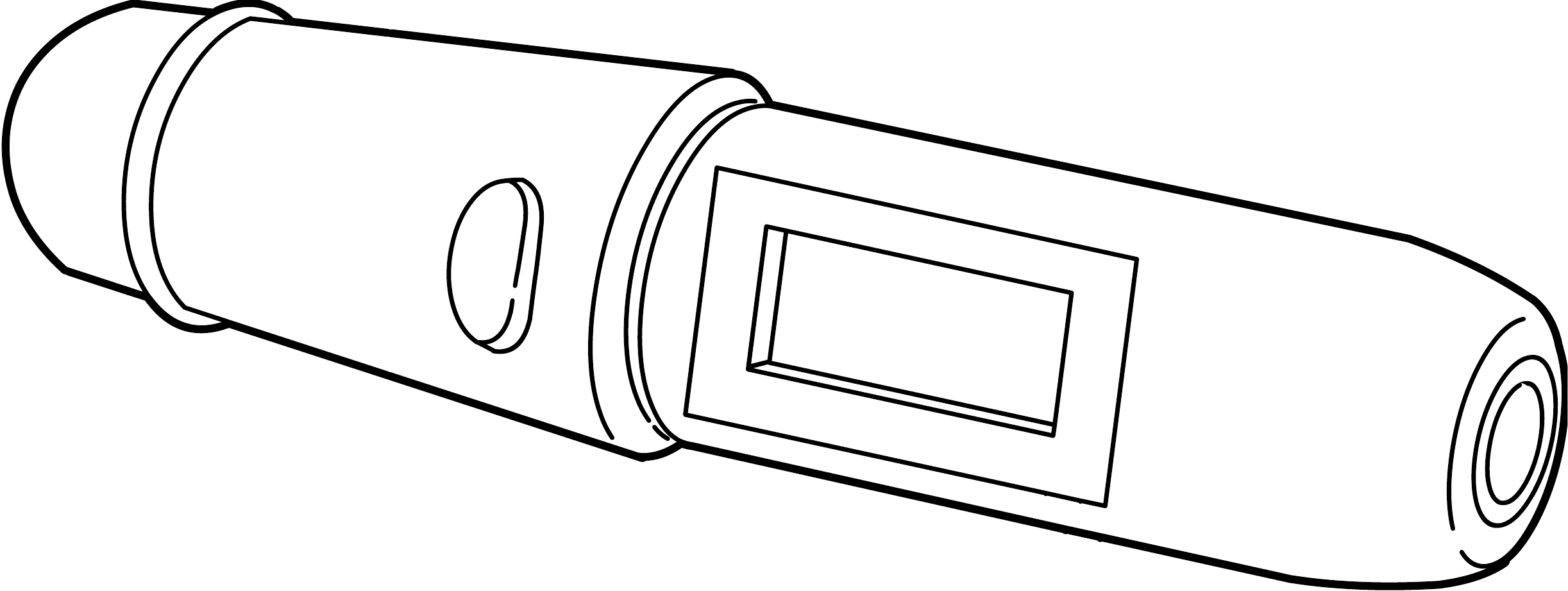}
        \caption*{DT8220 unit}
    \end{wrapfigure}
    Infrared thermometers detect the amount of thermal (or \emph{black-body}) radiation emitted by a warm object. It is hoped that we may be able to use this value to detect and distinguish the materials surrounding DORI, including trees, rocks, grass, and any living organisms that are present. We purchased a small, pen-style infrared thermometer from an online retailer, and connected small signal wires onto its display connector. We use the same screen sniffing technique detailed in "Construction - Nodes" to extract the displayed temperature reading, and this value is then output to the system bus for logging. The infrared thermometer is mounted on the sensor plate, aligned parallel to the laser distance meter.

    \subsubsection*{PIR motion}
We extracted a Passive Infrared (PIR) motion sensor from a broken home alarm system. This sensor is designed to detect the body heat radiated by living creatures, and it signals its output when a sudden change in heat is detected. We constantly monitor this signal in an attempt to photograph any wildlife that wanders near \textsc{DORI}.

    \subsubsection*{GPS}
    \begin{wrapfigure}{l}{0.19\textwidth}
        \centering
        \includegraphics[height=1.4cm]{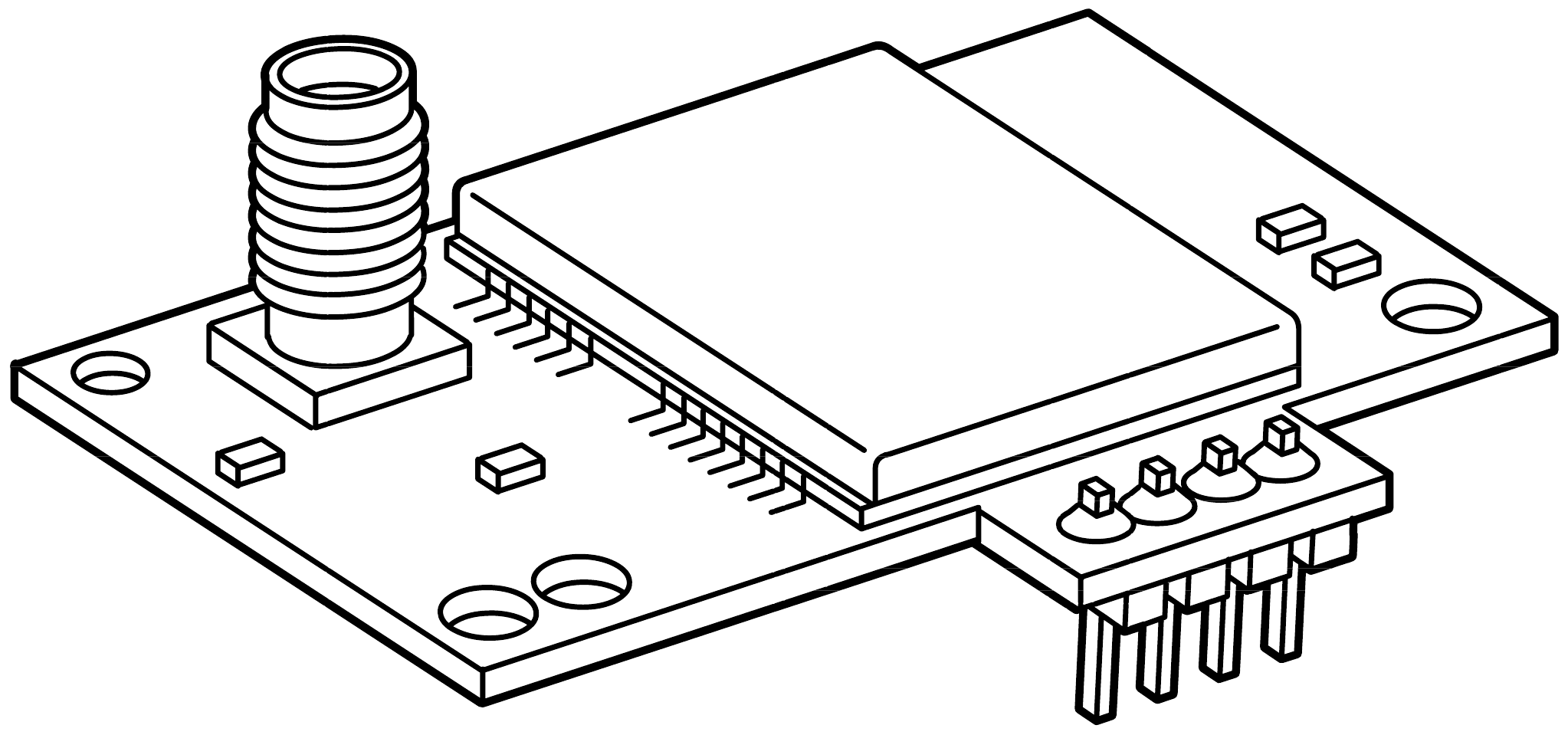}
        \caption*{LEA-4S module}
    \end{wrapfigure}
    Almost all GPS receivers output a variant of the NMEA 0183 standard which is a simple stream of ASCII sentences at 9600 baud containing plaintext descriptions of the unit's current location, speed, currently visible satellites, as well as the current time and date. This means a particular GPS driver will work with almost any commodity GPS receiver that includes an NMEA output pin. We were therefore relatively free to pick any off-the-shelf GPS receiver. DORI uses a \brand{uBlox LEA-4S} GPS module purchased from an electronics market in China. This unit includes a simple passive external antenna which we mounted on the top of the arm in order to maximize its received signal strength.

    The GPS node is configured to periodically output DORI's latitude and longitude after averaging several GPS sensor readings. This is because we don't expect DORI to move very much when we haven't sent any motor commands, so DORI's position should remain fixed between drive events. Every unique spot where DORI remains between drive events is referred to as a \emph{site}, and these are automatically logged and uniquely numbered by the \textsc{gateway} server when a drive command is sent. The operator can then use \textsc{tk}'s \emph{site editor} to manually reconstruct a particular site's surrounding environment based on the recorded sensor measurements.

    In addition to recording DORI's current position, the GPS node also extracts the current time and date from the NMEA output, and outputs a normalized unix timestamp onto the system bus for each node to synchronize to the same value. This limits the amount of clock drift that can occur as various nodes are outputting periodic sensor readings. Having this value available from the GPS satellites means that we don't need to include an additional RTC (real-time clock) hardware module. 

\subsection{Orientatation data}

    \subsubsection*{Accelerometer}
    \begin{wrapfigure}{l}{0.2\textwidth}
        \centering
        \includegraphics[height=4.5cm]{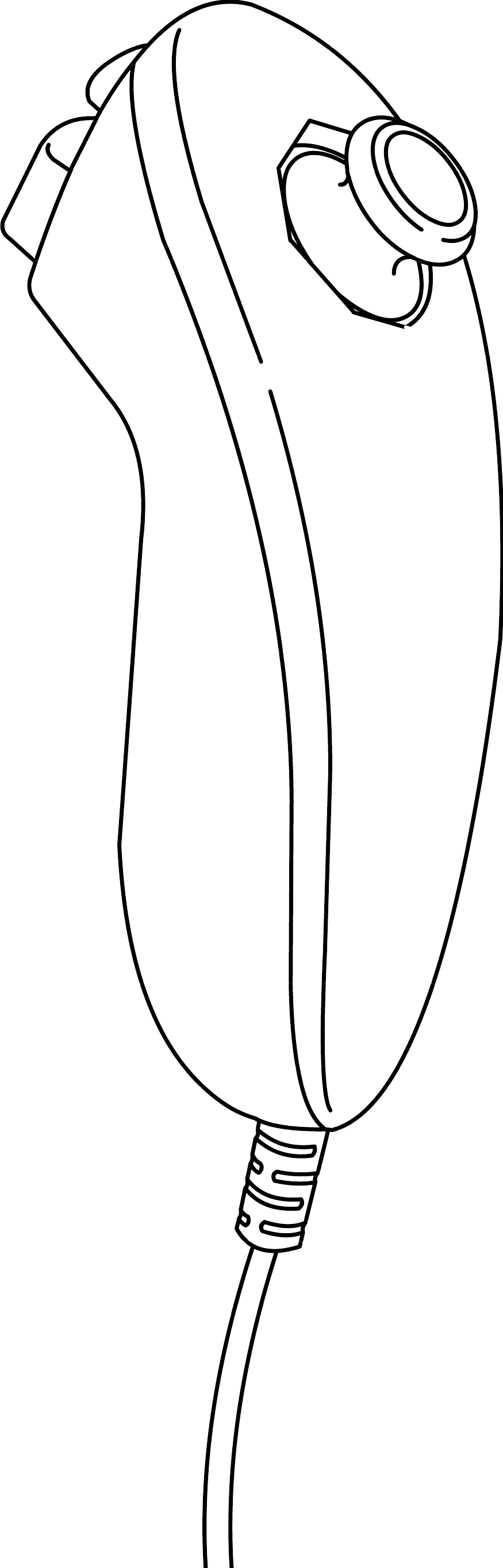}
        \caption*{Wii nunchuck}
    \end{wrapfigure}
    Linear motion accelerometers measure the sensor's linear acceleration along a number of perpendicular axes. For an object that is at rest on earth, the sensor's output will represent the acceleration due to gravity, and the direction of this vector can be used to calculate the sensor's physical orientation. 

    We chose to use \brand{Nintendo Wii} motion-sensitive controllers containing the \brand{Analog Devices ADXL335} 3D linear accelerometer to measure DORI's physical orientation. We found that these Nunchuck controllers are a very low-cost way to acquire the \brand{ADXL335} sensor along with all the necessary filtering capacitors and power regulation circuitry. Linear accelerometers are susceptible to drift, so we filter the measured values in order to obtain a more precise tilt measurement. 

    \subsubsection*{Magnetometer}
    DORI is also equipped with a \brand{Honeywell HMC5883L} 3-axis digital magnetometer. A magnetometer is a digital compass that can measure the magnetic field strength along several axes. These individual measurements can then be used to calculate a magnetic field vector representing a magnetic flux and its direction. Near Earth's equator the magnetic field lines run parallel to the ground, and the sensor's X and Y axis measurements can be directly used to calculate a bearing toward magnetic north. Nearer to the poles, the Earth's magnetic field lines run almost perpendicular to the ground, and a small amount of tilt will produce extremely large errors in the calculated bearing. Therefore in most applications an accelerometer's tilt measurement is combined with the magnetometer's X, Y, and Z axis values to produce an accurate tilt-compensated compass bearing~\cite{tilt}.

    Due to the large quantities of ferromagnetic metal used in the robot's construction, the magnetometer's output is subject to a large but predictable amount of magnetic bias (known as the \emph{soft-iron field}.) In order to measure the strength of the soft-iron field, an averaged magnetometer measurement was recorded in a large grass area free of large metal objects, and two more reference measurements were taken in the same position and orientation outside of DORI's chassis, with the sensor arm raised and lowered.

\begin{figure}[h]
    \scriptsize
    \centering
    \input{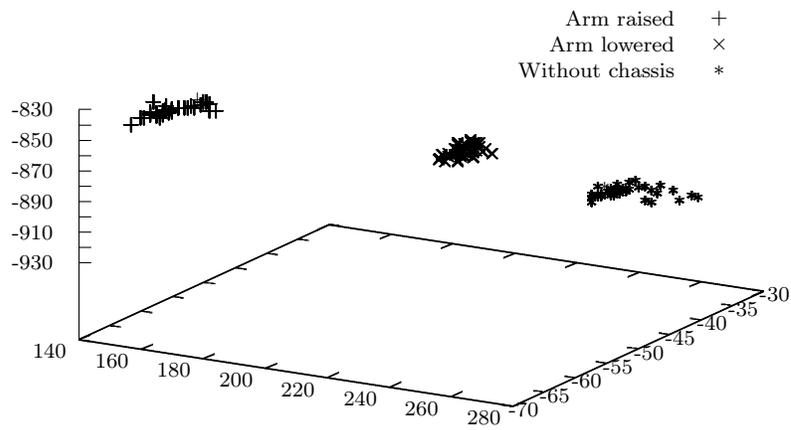}
    \caption{Magnetometer calibration for different arm positions}
\end{figure}
    
    The difference between these measurements represents the magnetic bias induced by the frame for each arm position, which must be subtracted from any subsequent measurements taken from within the metal frame. The state of DORI's arm is recorded by \textsc{gateway}, and these measurements are normalized by \textsc{gateway} before being stored for analysis. For magnetometer values recorded while the arm is between the raised and lowered positions, a bias estimate is created by linear interpolation between the two measured arm position bias values.

    \subsubsection*{Gyroscope}
    An \brand{InvenSense MPU-6050} digital gyroscope is also included in order to measure rotational acceleration along all 3 axes. This sensor is only activated during a drive event, in order to record a high precision motion profile of the terrain DORI is driving over. This 3-dimensional motion profile is then logged and transmitted back to \textsc{gateway}. The data will help reconstruct the terrain connecting the two sites that DORI has visited. The \brand{MPU-6050} unit also includes a 3-axis linear accelerometer, which can be used as a backup device in the event of a hardware failure in the primary accelerometer.

    \subsubsection*{Arm position feedback}
    The linear actuator used to raise and lower the arm has an integrated linear potentiometer which can be used as a voltage divider to produce a voltage range representing its current position. This value is connected to an AD (analog-to-digital) input pin of the motor control node, and can be used to raise and lower the arm to precise angles between 0$^\circ$ and 90$^\circ$ (fully raised.)

\section{Outputs}
    DORI's controllable outputs are all connected using a common interfacing technique. We made sure to pick devices that operated at 12v DC, and their power is controlled using mechanical relays. The list of devices which are connected in this way includes DC motors, lights, heating pads, Peltier thermoelectric cooling pads, as well as the stepper motor for the sensor plate and the linear actuator for raising the arm.

    The two DC motors, the linear actuator, and the thermoelectric cooling pads are all connected through simple H-bridge circuits made from two relays each. For the DC motors and the linear actuator, this gives us simple directional control using 2 output pins per device.

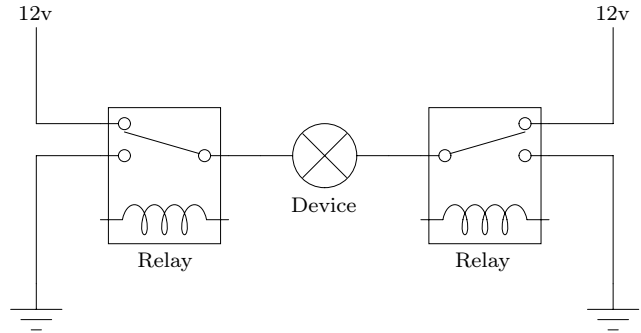
\begin{figure}[h]
    \centering
    \definecolor{cffffff}{RGB}{255,255,255}

\begin{tikzpicture}[y=0.80pt,x=0.80pt,yscale=-1, inner sep=0pt, outer sep=0pt]
\tikzstyle{every node}=[font=\scriptsize]
\begin{scope}[cm={{1.25,0.0,0.0,-1.25,(0.0,420.0)}}]
  \path[fill=cffffff,nonzero rule] (44.1985,287.5584) -- (86.6242,287.5584) --
    (86.6242,236.0428) -- (44.1985,236.0428) -- (44.1985,287.5584) -- cycle;
  \path[draw=black,line join=miter,line cap=round,miter limit=5.54,line
    width=0.297pt] (44.1985,287.5584) -- (86.6242,287.5584) -- (86.6242,236.0428)
    -- (44.1985,236.0428) -- (44.1985,287.5584) -- cycle(58.5930,245.1331) ..
    controls (57.0781,252.7095) and (51.0182,252.7095) ..
    (49.5017,245.1331)(58.5930,245.1331) .. controls (60.1080,237.5578) and
    (55.5631,237.5578) .. (57.0781,245.1331)(81.3210,245.1331) .. controls
    (79.8061,252.7095) and (73.7461,252.7095) ..
    (72.2312,245.1331)(73.7461,245.1331) .. controls (72.2312,252.7095) and
    (66.1694,252.7095) .. (64.6544,245.1331)(66.1694,245.1331) .. controls
    (64.6544,252.7095) and (58.5930,252.7095) ..
    (57.0781,245.1331)(73.7461,245.1331) .. controls (75.2611,237.5578) and
    (70.7147,237.5578) .. (72.2312,245.1331)(66.1694,245.1331) .. controls
    (67.6848,237.5578) and (63.1395,237.5578) ..
    (64.6544,245.1331)(41.1686,245.1331) -- (49.5017,245.1331)(89.6556,245.1331)
    -- (81.3210,245.1331);
  \path[draw=black,line join=miter,line cap=round,miter limit=5.54,line
    width=0.297pt] (78.6693,270.1339) -- (50.2599,278.3468);
  \path[draw=black,line join=miter,line cap=round,miter limit=5.54,line
    width=0.297pt] (78.2911,269.3772) .. controls (78.2911,270.6312) and
    (79.3087,271.6489) .. (80.5643,271.6489) .. controls (81.8195,271.6489) and
    (82.8375,270.6312) .. (82.8375,269.3772) .. controls (82.8375,268.1216) and
    (81.8195,267.1039) .. (80.5643,267.1039) .. controls (79.3087,267.1039) and
    (78.2911,268.1216) .. (78.2911,269.3772)(82.8375,269.3772) --
    (86.6242,269.3772);
  \path[draw=black,line join=miter,line cap=round,miter limit=5.54,line
    width=0.297pt] (52.5331,269.3772) .. controls (52.5331,270.6312) and
    (51.5155,271.6489) .. (50.2599,271.6489) .. controls (49.0044,271.6489) and
    (47.9867,270.6312) .. (47.9867,269.3772) .. controls (47.9867,268.1216) and
    (49.0044,267.1039) .. (50.2599,267.1039) .. controls (51.5155,267.1039) and
    (52.5331,268.1216) .. (52.5331,269.3772)(47.9867,269.3772) --
    (44.1985,269.3772)(89.6556,269.3772) -- (86.6242,269.3772)(41.1686,269.3772)
    -- (44.1985,269.3772)(52.5331,281.4985) .. controls (52.5331,280.2429) and
    (51.5155,279.2253) .. (50.2599,279.2253) .. controls (49.0044,279.2253) and
    (47.9867,280.2429) .. (47.9867,281.4985) .. controls (47.9867,282.7540) and
    (49.0044,283.7717) .. (50.2599,283.7717) .. controls (51.5155,283.7717) and
    (52.5331,282.7540) .. (52.5331,281.4985)(47.9867,281.4985) --
    (44.1985,281.4985)(41.1686,281.4985) -- (44.1985,281.4985);
  \path[fill=cffffff,nonzero rule] (165.4152,287.5584) -- (207.8405,287.5584) --
    (207.8405,236.0428) -- (165.4152,236.0428) -- (165.4152,287.5584) -- cycle;
  \path[draw=black,line join=miter,line cap=round,miter limit=5.54,line
    width=0.297pt] (165.4152,236.0428) -- (207.8405,236.0428) --
    (207.8405,287.5584) -- (165.4152,287.5584) -- (165.4152,236.0428) --
    cycle(193.4460,245.1331) .. controls (194.9610,252.7095) and
    (201.0224,252.7095) .. (202.5374,245.1331)(193.4460,245.1331) .. controls
    (191.9310,237.5578) and (196.4760,237.5578) ..
    (194.9610,245.1331)(170.7184,245.1331) .. controls (172.2334,252.7095) and
    (178.2948,252.7095) .. (179.8098,245.1331)(178.2948,245.1331) .. controls
    (179.8098,252.7095) and (185.8697,252.7095) ..
    (187.3861,245.1331)(185.8697,245.1331) .. controls (187.3861,252.7095) and
    (193.4460,252.7095) .. (194.9610,245.1331)(178.2948,245.1331) .. controls
    (176.7798,237.5578) and (181.3247,237.5578) ..
    (179.8098,245.1331)(185.8697,245.1331) .. controls (184.3547,237.5578) and
    (188.9011,237.5578) .. (187.3861,245.1331)(210.8705,245.1331) --
    (202.5374,245.1331)(162.3853,245.1331) -- (170.7184,245.1331);
  \path[draw=black,line join=miter,line cap=round,miter limit=5.54,line
    width=0.297pt] (173.3698,270.1339) -- (201.7791,278.3468);
  \path[draw=black,line join=miter,line cap=round,miter limit=5.54,line
    width=0.297pt] (173.7484,269.3772) .. controls (173.7484,270.6312) and
    (172.7318,271.6489) .. (171.4766,271.6489) .. controls (170.2196,271.6489) and
    (169.2034,270.6312) .. (169.2034,269.3772) .. controls (169.2034,268.1216) and
    (170.2196,267.1039) .. (171.4766,267.1039) .. controls (172.7318,267.1039) and
    (173.7484,268.1216) .. (173.7484,269.3772)(169.2034,269.3772) --
    (165.4152,269.3772);
  \path[draw=black,line join=miter,line cap=round,miter limit=5.54,line
    width=0.297pt] (199.5074,269.3772) .. controls (199.5074,270.6312) and
    (200.5251,271.6489) .. (201.7791,271.6489) .. controls (203.0347,271.6489) and
    (204.0523,270.6312) .. (204.0523,269.3772) .. controls (204.0523,268.1216) and
    (203.0347,267.1039) .. (201.7791,267.1039) .. controls (200.5251,267.1039) and
    (199.5074,268.1216) .. (199.5074,269.3772)(204.0523,269.3772) --
    (207.8405,269.3772)(162.3853,269.3772) --
    (165.4152,269.3772)(210.8705,269.3772) --
    (207.8405,269.3772)(199.5074,281.4985) .. controls (199.5074,280.2429) and
    (200.5251,279.2253) .. (201.7791,279.2253) .. controls (203.0347,279.2253) and
    (204.0524,280.2429) .. (204.0524,281.4985) .. controls (204.0524,282.7540) and
    (203.0347,283.7717) .. (201.7791,283.7717) .. controls (200.5251,283.7717) and
    (199.5074,282.7540) .. (199.5074,281.4985)(204.0524,281.4985) --
    (207.8405,281.4985)(210.8705,281.4985) -- (207.8405,281.4985);
  \path[fill=cffffff,nonzero rule] (126.0195,257.2559) .. controls
    (132.7144,257.2559) and (138.1408,262.6822) .. (138.1408,269.3772) .. controls
    (138.1408,276.0706) and (132.7144,281.4985) .. (126.0195,281.4985) .. controls
    (119.3261,281.4985) and (113.8982,276.0706) .. (113.8982,269.3772) .. controls
    (113.8982,262.6822) and (119.3261,257.2559) .. (126.0195,257.2559);
  \path[draw=black,line join=miter,line cap=round,miter limit=5.54,line
    width=0.297pt] (126.0195,257.2559) .. controls (132.7144,257.2559) and
    (138.1408,262.6822) .. (138.1408,269.3772) .. controls (138.1408,276.0706) and
    (132.7144,281.4985) .. (126.0195,281.4985) .. controls (119.3261,281.4985) and
    (113.8982,276.0706) .. (113.8982,269.3772) .. controls (113.8982,262.6822) and
    (119.3261,257.2559) .. (126.0195,257.2559)(150.2636,269.3772) --
    (138.1408,269.3772)(101.7769,269.3772) --
    (113.8982,269.3772)(117.7692,277.6275) --
    (134.2717,261.1250)(134.2717,277.6275) -- (117.7692,261.1250);
  \path[draw=black,line join=miter,line cap=round,miter limit=5.54,line
    width=0.297pt] (16.9260,281.4985) -- (41.1686,281.4985)(16.9260,269.3772) --
    (16.9260,220.8901)(16.9260,269.3772) -- (41.1686,269.3772)(210.8705,269.3772)
    -- (235.1146,269.3772)(235.1146,269.3772) -- (235.1146,220.8901);
  \path[draw=black,line join=miter,line cap=round,miter limit=5.54,line
    width=0.297pt] (16.9260,211.1927) -- (16.9260,220.8901)(26.6234,211.1927) --
    (7.2286,211.1927)(22.7439,207.3147) -- (11.1066,207.3147)(18.8655,203.4352) --
    (14.9861,203.4352)(235.1146,211.1927) --
    (235.1146,220.8901)(244.8120,211.1927) --
    (225.4175,211.1927)(240.9325,207.3147) --
    (229.2955,207.3147)(237.0530,203.4352) -- (233.1750,203.4352);
  \path[draw=black,line join=miter,line cap=round,miter limit=5.54,line
    width=0.297pt] (16.9260,317.8624) -- (16.9260,281.4985)(210.8705,281.4985) --
    (235.1146,281.4985)(235.1146,317.8624) -- (235.1146,281.4985);
  \path[draw=black,line join=miter,line cap=round,miter limit=5.54,line
    width=0.297pt] (89.6556,269.3772) -- (101.7769,269.3772)(150.2636,269.3772) --
    (162.3853,269.3772);
  \path[xscale=1.000,yscale=-1.000,fill=black] (10.048147,-320.72159) node[above
    right] (text3109) {12v};
  \path[xscale=1.000,yscale=-1.000,fill=black] (228.40594,-320.78854) node[above
    right] (text3113) {12v};
  \path[xscale=1.000,yscale=-1.000,fill=black] (113.36832,-248.19261) node[above
    right] (text3117) {Device};
  \path[xscale=1.000,yscale=-1.000,fill=black] (55.147778,-225.77431) node[above
    right] (text3121) {Relay};
  \path[xscale=1.000,yscale=-1.000,fill=black] (175.32773,-225.85657) node[above
    right] (text3125) {Relay};
\end{scope}

\end{tikzpicture}
    \caption[H-bridge circuit]{A simple H-bridge using 2 relays (not all connections shown)}
\end{figure}

    The thermoelectric cooling pad uses the Peltier effect to create a temperature differential when an electric current passes through a bond joining two types of metal -- one side of the bond will be heated, and the other side will be cooled. When the direction of the current is reversed, the direction of the heat differential is reversed and the temperature of the two sides is swapped. We have attached these cooling pads on our two most temperature-sensitive devices, the camera and the laser range finder. In this way we can carefully regulate the temperature of these devices, and when the temperature falls outside the acceptable range the device can be cooled or heated as required.

    The lights, heating pads, and the battery contingency circuit (see \emph{Battery level sensor}) are connected through a single relay, giving us simple on-off control.

\section{Auxiliary}
    \subsubsection*{Data logger}
    During normal operation DORI's nodes are configured to periodically capture sensor measurements and broadcast them on the main system bus. These measurements are automatically recorded by the logger node, which writes all system bus traffic to an SD card to be transferred back to \textsc{gateway}. The only system bus traffic which is excluded from the data log is file transfer packets, because these contain data which are already stored elsewhere.

    This operation essentially represents the core activity during DORI's \emph{passive} operating mode, and is therefore more important to DORI's operation than most other nodes. DORI is thus equipped with an alternative logging strategy in the event that the data logger node fails. The node which controls the sensor plate electronics including the laser range finder, the infrared thermometer, and the Powershot digital camera is also able to access the camera's internal SD card. In the event of a hardware failure of the data logger node, we are therefore able to reflash the sensor plate node with an alternative firmware that includes data logging capabilities, and continue normal operation.

    The function of the logger node is critical for data collection during DORI's passive operating mode, as all sensors broadcast onto the system bus are expected to be picked up by the logger for permanent storage. In the event of a hardware failure on the logger node, the node used to control the Powershot camera can be used as a replacement logger node, and can be reflashed to log all system bus traffic to the Powershot's internal SD card.

    \subsubsection*{Image capture}
    As a simple image capture solution we decided to modify a consumer-grade \brand{Canon Powershot A540} to be controlled by a microcontroller. By soldering onto the buttons and switches of the camera, we were able to provide an interface for the AVR to take pictures, navigate menus, and power cycle the device. The \brand{Powershot} saves the captured images, video, and sound clips to an attached SD card, and the AVR can use its own wires connected to the same SD card to read the resulting file.

    \subsubsection*{Realtime clock}
    The clocks used on DORI's internal nodes can incur up to several minutes of clock drift, depending on several factors including temperature, physical orientation, and gradual aging effects. This clock source can be used for simple timers used to output periodic measurements from various sensors, but there will be an unpredictable amount of drift in the timestamps of the reported measurements. One solution would be to remotely send a command to update the absolute timestamp stored on each node at periodic intervals, however this would require periodic connections just for this purpose.

    We had originally planned to include a \brand{Maxim 1-Wire DS1904} RTC (real-time clock) module to periodically update every node's internal timestamp from its high-quality internal clock source. The \brand{DS1904} module is internally temperature-compensated and contains its own internal battery rated for a 10-year lifetime. This provides a stable independent clock source with minimal extra hardware complexity, as a compatible 1-Wire network had already been implemented for the \brand{DS18B20} temperature sensors.

    Another candidate clock source was the \brand{WWVB} broadcast signal, which is used to synchronize millions of low-cost radio controlled clocks in North America. This signal can be picked up using a small circuit, or the signal could be directly read from a commodity radio controlled clock unit.

    Eventually we decided to directly use the clock signal received by the GPS node which is generated by the high-precision atomic clocks within the GPS satellites. This introduced a new problem, as DORI's internal timestamp values were now more accurate than the system clock of the \textsc{gateway} server. We found the GPS timestamp to be virtually identical to the value broadcast by NASA's NTP service at \texttt{ntp.time.gov}, so the \textsc{gateway} server was configured to synchronize its system clock over NTP every day.

    \subsubsection*{Solar panel}
    DORI has two separate batteries and these are used to power the high power devices and logic devices, respectively. We purchased two solar panels from an online retailer meant for trickle-charging a vehicle battery, and these are used to provide power once DORI is deployed. The solar panels charge both batteries to a nominal 12.7v, and have internal charge controllers to protect the batteries against overcharging and overvoltage.

    The solar panels are mounted on DORI's arm in order to avoid accumulation of leaves and other debris. If a large amount of material has accumulated on the solar panels, the operator can raise the arm to a 45$^{\circ}$ angle and wait for incoming rainfall to dislodge the material.

    \subsubsection*{Battery level sensor}
    During DORI's normal operation it is critical that the battery power is never allowed to fall below 11v or so, as this could interfere with the operator's ability to communicate with DORI and retrieve sensor measurements. If the power level were to fall below a usable level, DORI's nodes would immediately detect the brown-out condition and attempt to send a final signal before powering off. As the solar panels gradually recharge the batteries,  DORI's nodes would power on and normal modem communications can resume. This hibernation state allows for continued operation even during periods of extremely low solar panel output. The voltages of the two batteries are fed into a sensor node as inputs allowing the operator to track the battery charge levels over time. If the charge level is found to be too low, one or more of DORI's sensor nodes can be put into a deep sleep state to minimize power use until the batteries have been recharged.

    Despite being electrically isolated, the batteries are able to be bridged together using a contingency circuit consisting of two relays and a low-value resistor. This will only be used in an emergency situation if the battery used for the logic circuitry or its solar panel experiences a malfunction and is no longer able to power the main system electronics. The two batteries will then be bridged to allow the failed battery to be supported by the other functioning battery.

    \subsubsection*{Mirror}
    A small convex stainless steel mirror is mounted on DORI's arm to one side of the rotating sensor plate. During operation, the Powershot camera can be aimed at this mirror to capture an image of DORI's main chassis. This can be used to assess DORI's physical condition in the event of a physical disturbance or other obstruction that has interfered with DORI's operation. Having a mirror means that we are able to capture images of various angles that would not otherwise be possible without adding an additional degree of freedom to the movement of the arm or sensor plate.

    \subsubsection*{Patching}
In the event that a software bug is found after DORI has been deployed to a remote location, we've written a simple patching procedure that can be executed to update the software running on any of the nodes. 
The patching procedure is performed in two steps. We begin by transmitting the updated firmware image to one of the SD cards (note that both the logger and the Powershot camera have SD cards that can be used for this purpose.) Once the entire firmware image has been uploaded, DORI computes a CRC-16 checksum to verify that the file wasn't corrupted during transmission. If there are no errors with the computed checksum, then the target node can be reflashed. A system bus command is sent to the node to send it into a reflashing loop, and the contents of the firmware image file are streamed over the system bus. Chunks are flashed on the fly, and the target node replies with a \emph{Clear To Send} (CTS) message after every file chunk for strict flow control. Once the node's firmware has been fully reflashed, it begins executing the new code.

\chapter{Conclusion}
    We have designed and built a simple reliable robot for remote exploration and data collection. Over the next several months we will continue to test DORI's various systems for faults. DORI will be tested for mechanical and electrical equipment failure, as well as complete communication loss. After DORI's failsafe systems have been fully tested under various weather conditions, we will permanently deploy DORI to a remote location in order to prove the system's reliability.

        \section{Future development}
    During DORI's development we found several aspects of the design that could be improved, including the cellular-based communication link and the sensor plate mechanism. These components will be kept as-is during DORI's testing. We will research alternate solutions if any other issues are found.
    
    Several additional peripherals were originally planned but were removed from the final design for various reasons. These include a satellite phone for communications, a piezoelectric disdrometer to measure rain drop distribution, and a physical odometer to provide feedback on wheel rotations. These devices will likely be added to a future robot design.
    
    \section{Results}
    After preliminary testing, we have found DORI's distributed archicture to be very reliable and simple to develop. Any number of sensor nodes could be added to the final design without changing the system architecture. The design's simplicity is due to the small amount of data gathered -- a faster processor archicture would be needed to handle greater quantities of data. However, as a simple data collection system, DORI's distributed node design performs within our expectations.

    We are currently collecting performance and reliability data in order to test DORI's failure response systems. We intend to carry out 8-10 months of further testing, and we expect to fully deploy DORI in April 2014.

\begin{appendices}
\chapter{Bill of materials}
\index{appendices}

\begin{longtable}{ |l|l|l| }

  \hline
  Base frame & Scrap streetlight enclosure & \$0 \\
  Wheels & Scrap tractor wheels & \$20 \\
  Arm & Scrap floor jack & \$0 \\
  Linear actuator & aliexpress.com & \$30 \\
  Drive motors & Scrap BMW power window motors & \$30 \\
  Drive shafts & Scrap lawnmower drive shafts & \$0 \\
  Chains & Princess Auto & \$30 \\
  Sensor plate & Scrap aluminum & \$0 \\
  Stepper motor & aliexpress.com & \$25 \\
  BMP085 & aliexpress.com & \$8 \\
  Wii Nunchuck & aliexpress.com & \$8 $\times$ 2 \\
  Heating pads & aliexpress.com & \$10 \\
  Peltier pads & aliexpress.com & \$10 \\
  Canon Powershot A540 & craigslist.org & \$10 \\
  ublox LEA-4S GPS module & aliexpress.com & \$25 \\
  WH-0531 wireless rain gauge & ebay.ca & \$20 \\
  ATmega88 & taobao.com & \$2 $\times$ 10 \\
  BenQ m32 GSM modem & aliexpress.com & \$16 $\times$ 2 \\
  GSM antenna & dx.com & \$10 \\
  Nokia 3390 cellphone & craigslist.org & \$15 \\
  Rubber sealant & Home Depot & \$10 \\
  Metal armored cable & Scrap from junkyard & \$0 \\
  DS19B20 temperature sensor & microchip.com free sample & \$0 $\times$ 12 \\
  4GB SD card & dx.com & \$5 $\times$ 2 \\
  MCP2515 CAN controller & microchip.com free sample & \$0 $\times$ 16 \\
  MCP2551 CAN transceiver & microchip.com free sample & \$0 $\times$ 16 \\
  Passive IR sensor & Scrap motion detector & \$0 \\
  HC-SR04 ultrasonic distance sensor & aliexpress.com & \$3 \\
  Litmus paper & aliexpress.com & \$2 \\
  Plastic tube, foam pipe & Home Depot & \$5 \\
  Waterproof ultrasonic transducer & Salvaged from broken car bumper & \$0 \\
  Infrared thermometer & aliexpress.com & \$9 \\
  LDM-40 laser distance meter & aliexpress.com & \$65 \\  
  Relays & Lee's Electronics & \$30 \\
  Flood light & Princess Auto & \$20 \\
  Convex mirror & Princess Auto & \$10 \\
  MQ-2 & aliexpress.com & \$1.50 \\
  10W solar panel & aliexpress.com & \$35 \\

\hline
\end{longtable}

\end{appendices}

\nocite{*}     
\renewcommand{\baselinestretch}{\tighttextstretch} 
\normalsize
\bibliographystyle{plain}   
\addcontentsline{toc}{chapter}{Bibliography}
\typeout{Bibliography}
\begin{CJK*}{UTF8}{gbsn}
\bibliography{draft}
\end{CJK*}
\renewcommand{\baselinestretch}{\textstretch} 
\normalsize

\end{document}